\documentclass{article}
\usepackage{iclr2027_conference,times}

\usepackage{amsmath,amsfonts,bm}

\def\eqref#1{equation~\ref{#1}}

\def\1{\bm{1}}

\DeclareMathAlphabet{\mathsfit}{\encodingdefault}{\sfdefault}{m}{sl}
\SetMathAlphabet{\mathsfit}{bold}{\encodingdefault}{\sfdefault}{bx}{n}

\usepackage{amsmath,amssymb}
\usepackage{graphicx}
\usepackage[hypertexnames=false]{hyperref}
\usepackage{algorithm}
\usepackage{booktabs}
\usepackage{float}
\usepackage{wrapfig}
\usepackage{needspace}
\usepackage{listings}
\usepackage{placeins}
\usepackage{tabularx}
\usepackage{url}
\usepackage{booktabs,multirow}

\definecolor{joltteacher}{HTML}{9A4F2F}
\definecolor{joltstudent}{HTML}{286C96}

\title{A Good Self-Teacher Meets the Student Where They Are: Joint On-Policy Learning and Teaching}

\author{%
Randy Ardywibowo\\
Perplexity\\
{\small\href{mailto:randy@perplexity.ai}{\texttt{randy@perplexity.ai}}}
\And
Arnav Dalal\\
Perplexity\\
{\small\href{mailto:arnav.dalal@perplexity.ai}{\texttt{arnav.dalal@perplexity.ai}}}
\And
Jiantao Jiao\\
NVIDIA\\
{\small\href{mailto:jiantaoj@nvidia.com}{\texttt{jiantaoj@nvidia.com}}}
}

\hypersetup{
  pdftitle={A Good Self-Teacher Meets the Student Where They Are: Joint On-Policy Learning and Teaching},
  pdfauthor={Randy Ardywibowo; Arnav Dalal; Jiantao Jiao},
  pdfsubject={Joint on-policy learning and teaching for language models},
  pdfkeywords={self-distillation, reinforcement learning, privileged information, on-policy learning},
  hidelinks
}

\iclrfinalcopy
\begin{document}
\raggedbottom

\maketitle
\lhead{}

\begin{abstract}
Reinforcement Learning (RL) from outcome rewards suffers from sparse supervision, particularly on difficult, long-horizon tasks where successful trajectories are rare and costly to generate. On-Policy Distillation (OPD) offers an attractive alternative by providing dense token-level supervision from a stronger teacher along the student’s own generations. Self-distillation methods further remove the need for a separate teacher model by conditioning the same policy on privileged information to serve as its own teacher. However, privileged conditioning alone does not guarantee that the resulting distillation update improves the student. Indeed, privileged information can lead the teacher to solve tasks through shortcuts unavailable to the student, producing supervision poorly matched to the student’s current behavior. Consequently, even a higher-performing teacher can provide guidance that degrades student performance. To address this, we analyze how the choice of privileged teacher affects the student’s update. We derive a necessary and sufficient condition for the teacher’s local distillation update to be a positive multiple of the student’s reward gradient. Our analysis suggests that the teacher should not only perform well on the task, but also provide guidance suited to the student’s current capabilities. This characterization motivates a practical teacher-training surrogate that combines outcome rewards with token-level Kullback–Leibler (KL) regularization toward the student. Based on this result, we propose Joint On-Policy Learning and Teaching (JOLT), which jointly trains a single policy in two roles: a privileged teacher using a KL-regularized objective, and an unprivileged student using dense on-policy distillation. Our experiments across mathematical reasoning, code generation, tool use, and agentic terminal-use tasks demonstrate improvements in both training efficiency and evaluation performance. On MATH, JOLT uses $13\times$ fewer completion tokens and $6.5\times$ fewer processed training tokens than GRPO at matched accuracy, without direct student reward. With student outcome rewards, JOLT+ achieves absolute gains over GRPO of +4.2\% on MATH, +3.0\% on LCBv6, +9.9\% on AppWorld, and +4.5\% in Terminal-Bench pass@32.
\end{abstract}
\suppressfloats[t]

\begingroup
\setlength{\parskip}{6pt}
\section{Introduction}

\begin{figure}[t]
\centering
\begin{minipage}[c]{0.39\linewidth}
\includegraphics[width=\linewidth]{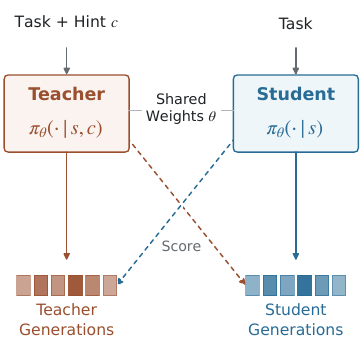}
\end{minipage}%
\hspace{0.02\linewidth}%
\begin{minipage}[c]{0.59\linewidth}
\par\hrule height 0.8pt\kern2pt
\begin{lstlisting}[
language=Python,
basicstyle=\fontencoding{T1}\fontfamily{pcr}\fontsize{7pt}{8.2pt}\selectfont,
commentstyle=\color{gray},
keywordstyle=\mdseries,
columns=fullflexible,
keepspaces=true,
showstringspaces=false,
breaklines=false,
emph={teacher,teacher_lp,beta_T},
emphstyle=\color{joltteacher},
emph={[2]student,student_lp,beta_S},
emphstyle={[2]\color{joltstudent}},
aboveskip=0pt,
belowskip=0pt
]
# Joint Training Step
# Shared Policy (hint: Privileged Information c)
teacher = policy.generate([task, hint])
student = policy.generate([task])

# Cross-Context Scores
student_lp = policy.score([task], teacher.tokens)
teacher_lp = policy.score([task, hint], student.tokens)

teacher.adv = student.adv = 0
teacher.adv += advantage(teacher)
teacher.adv += beta_T * (student_lp - teacher.logprobs)
student.adv += beta_S * (teacher_lp - student.logprobs)
student.adv += advantage(student)  # JOLT+ Only

policy.update([teacher, student])
\end{lstlisting}
\par\kern2pt\hrule
\end{minipage}
\par
\setlength{\abovecaptionskip}{5pt}
\caption{Joint On-Policy Learning and Teaching (JOLT) trains a shared model
as both a privileged teacher and an unprivileged student. The teacher learns
from outcome rewards with KL regularization toward the student, while the
student receives teacher supervision along its own generations. (Left): the
joint teacher--student training process. (Right):
pseudocode for a training step.}
\label{fig:jolt-overview}
\vspace{-6pt}
\end{figure}

Reinforcement Learning (RL) has become an effective way to improve language
models on reasoning tasks
\citep{shao2024deepseekmath,openai2024o1,deepseekai2025r1}. Many such methods
rely on outcome rewards observed only after a trajectory ends,
resulting in a sparse learning signal. This sparsity is especially challenging
for difficult, long-horizon agentic tasks, where successful trajectories may be
rare and costly to generate
\citep{yang2026programbench,li2026longhorizonterminalbench}.

On the other hand, On-Policy Distillation (OPD) provides a complementary source
of dense supervision. Rather than learning only from a final reward, OPD samples
trajectories from the student and evaluates each sampled action under a stronger
teacher. The teacher-student log-probability ratio then provides a dense
learning signal at every visited state \citep{agarwal2024onpolicy}. More
recently, On-Policy Self-Distillation (OPSD) removes the need for a separate
teacher model. The student policy serves as its own teacher given privileged information,
such as a verified solution, feedback, or a reference trace
\citep{zhao2026selfdistilled,penaloza2026privileged}.

Despite growing interest in self-distillation methods that use privileged
information, privileged conditioning alone does not guarantee that the
resulting distillation update improves the student
\citep{chakraborty2026pedagogicalrl,kaur2026rethinking}. Privileged
information can lead the teacher to exploit shortcuts unavailable to the student,
producing supervision poorly matched to the student's current behavior.
The teacher may also become unreliable
when continuing from an erroneous student-generated prefix. A privileged teacher can therefore achieve
higher expected reward while inducing an OPD update that locally decreases the
student's expected reward.

Teacher quality is therefore not determined by privileged task performance
alone, but by the learning signal provided to the current student. We derive a
necessary and sufficient condition for the teacher's local distillation update
to be a positive multiple of the student's reward gradient. A sufficient pointwise construction reveals the
principle underlying our approach: a good teacher meets the student where they
are. The ideal teacher favors high-advantage actions under student continuations
while remaining close to the student's policy. This motivates a practical
teacher-training surrogate combining outcome rewards with a teacher-student
Kullback--Leibler (KL) penalty.

Because the resulting teacher is defined relative to the current student,
learning and teaching are inherently coupled: as the student changes, so does
the teacher best suited to it. Joint On-Policy Learning and Teaching (JOLT)
closes this loop by jointly training both roles of the same policy, as illustrated in
Figure~\ref{fig:jolt-overview}. The privileged teacher learns from outcome rewards
with token-level KL regularization toward the student, while the unprivileged
student learns from the teacher's dense distillation signal. Each role learns
from trajectories generated under its own context, keeping both updates
on-policy, and both update the shared parameters. When student outcome rewards
are available, JOLT+ supplements the student's distillation update with direct
reinforcement learning.

Specifically, we make the following contributions:

\begingroup
\setlength{\leftmargini}{1.5em}
\begin{itemize}
    \item We characterize what makes a privileged teacher useful to
    its student. Higher teacher performance alone does not ensure that
    distillation improves the student. We derive a necessary and sufficient
    condition for the teacher's local distillation update to follow the
    student's reward gradient. This characterization motivates training the
    teacher to perform well while remaining close to the student's current
    policy.

    \item We introduce JOLT, which learns the teacher alongside the
    student rather than simply imitating a privileged policy. A single model
    serves both roles: the teacher learns from rewards with teacher--student
    KL regularization, while the student receives dense distillation
    supervision. Both roles learn on-policy from their own generations.
    JOLT+ additionally incorporates student outcome rewards.

    \item We demonstrate that this approach improves performance
    across reasoning and agentic tasks. Across mathematical reasoning, code
    generation, tool use, and terminal interaction, JOLT+ achieves absolute
    gains over GRPO of 4.2 percentage points on MATH, 3.0 on LCBv6, 9.9 on
    AppWorld, and 4.5 in Terminal-Bench pass@32.

    \item We show that strong performance can be reached with
    substantially less training. On MATH, JOLT matches GRPO's accuracy using
    $13\times$ fewer completion tokens and $6.5\times$ fewer processed
    training tokens, without direct rewards on student trajectories.
\end{itemize}
\endgroup

\endgroup
\begingroup
\setlength{\textfloatsep}{10pt plus 2pt minus 2pt}
\setlength{\floatsep}{10pt plus 2pt minus 2pt}
\setlength{\abovedisplayskip}{4pt plus 1pt minus 2pt}
\setlength{\belowdisplayskip}{4pt plus 1pt minus 2pt}
\setlength{\abovedisplayshortskip}{0pt plus 1pt}
\setlength{\belowdisplayshortskip}{2pt plus 1pt minus 1pt}
\section{Methodology}
\vspace{-3pt}

\subsection{Background}
\vspace{-2pt}

\noindent\textbf{Reinforcement Learning.}
We consider a language-model policy \(\pi_\theta\) acting in an episodic
decision process. At each step \(t\), the policy observes a state \(s_t\) and
samples an action \(a_t\). The resulting sequence forms a trajectory
\(\tau=(s_0,a_0,s_1,a_1,\ldots,s_T)\).
Rolling out the policy induces a trajectory distribution, denoted
\(\tau\sim\pi_\theta\).
Let \(R(\tau)\in\mathbb{R}\) denote the return assigned to trajectory \(\tau\).
The expected-return objective is
\(\mathcal{J}_{\mathrm{RL}}(\theta)
=\mathbb{E}_{\tau\sim\pi_\theta}[R(\tau)]\).
The corresponding policy gradient for this objective can be expressed using the
score-function estimator \citep{williams1992simple,sutton2000policy}:
\begin{equation}
\nabla_\theta \mathcal{J}_{\mathrm{RL}}(\theta)
=
\mathbb{E}_{\tau\sim\pi_\theta}
\left[
\sum_t
R(\tau)
\nabla_\theta\log\pi_\theta(a_t\mid s_t)
\right].
\end{equation}
Although unbiased, this estimator can have high variance because the gradient
contribution associated with \(a_t\) varies with subsequent actions and other
trajectory randomness. Conditioning the return on the current state and action
averages over this future randomness while preserving the expected policy
gradient, with
\(Q^{\pi_\theta}(s_t,a_t)=\mathbb{E}[R(\tau)\mid s_t,a_t]\).
The gradient also remains unchanged if a state-dependent baseline \(b(s_t)\) is
subtracted from the action value. Since the baseline does not depend on the
sampled action,
\begin{equation}
\mathbb{E}_{a_t\sim\pi_\theta(\cdot\mid s_t)}
\left[
b(s_t)
\nabla_\theta\log\pi_\theta(a_t\mid s_t)
\right]
=
0.
\end{equation}
The subtracted term therefore acts as a zero-mean control variate, allowing an
appropriate baseline to reduce variance without biasing the policy gradient.
Taking the baseline to be the expected action value under the current policy
centers each action value relative to the alternatives available at the same
state. This yields the on-policy advantage, with
\begin{equation}
V^{\pi_\theta}(s_t)
=\mathbb{E}_{a\sim\pi_\theta(\cdot\mid s_t)}[Q^{\pi_\theta}(s_t,a)],
\qquad
A^{\pi_\theta}(s_t,a_t)
=Q^{\pi_\theta}(s_t,a_t)-V^{\pi_\theta}(s_t).
\end{equation}
The policy gradient can therefore be written as
\begin{equation}
\nabla_\theta \mathcal{J}_{\mathrm{RL}}(\theta)
=
\mathbb{E}_{\tau\sim\pi_\theta}
\left[
\sum_t
A^{\pi_\theta}(s_t,a_t)
\nabla_\theta\log\pi_\theta(a_t\mid s_t)
\right].
\end{equation}
Thus, \(A^{\pi_\theta}(s_t,a_t)\) measures the expected excess return of
\(a_t\) relative to the current policy at \(s_t\). Using it as the action-level
weight exactly recovers the policy gradient of expected return. Since computing
\(Q^{\pi_\theta}\) and \(V^{\pi_\theta}\) requires expectations over future
trajectories, \(A^{\pi_\theta}\) is generally unavailable and must be estimated
from sampled returns or a learned value function
\citep{schulman2016highdimensional}.
In particular, when the return is observed only through a sparse terminal
reward, estimating useful action-level credit may require many rollouts.

\noindent\textbf{On-Policy Distillation.}
On-policy distillation (OPD) instead obtains dense supervision from a stronger
teacher policy \(\pi_T\) \citep{agarwal2024onpolicy}. Let \(\bar\theta\) denote
the parameters used to collect trajectories. OPD minimizes the reverse KL at
student-generated prefixes, holding the sampled prefix distribution and teacher
targets fixed during differentiation:
\begin{equation}
\mathcal{J}_{\mathrm{OPD}}(\theta;\bar\theta)
=
-
\mathbb{E}_{\tau\sim\pi_{\bar\theta}}
\left[
\sum_t
\mathrm{KL}
\left(
\pi_\theta(\cdot\mid s_t)
\;\middle\|\;
\pi_T(\cdot\mid s_t)
\right)
\right].
\end{equation}
The local gradient with respect to \(\theta\), evaluated at \(\theta=\bar\theta\), is
\begin{equation}
g_{\mathrm{OPD}}(\bar\theta;\pi_T)
=
\mathbb{E}_{\tau\sim\pi_{\bar\theta}}
\left[
\sum_t
\left(
\log\pi_T(a_t\mid s_t)
-
\log\pi_{\bar\theta}(a_t\mid s_t)
\right)
\nabla_{\bar\theta}
\log\pi_{\bar\theta}(a_t\mid s_t)
\right].
\end{equation}
The teacher-student log-probability ratio therefore supervises each sampled
action, reducing reliance on sparse rewards but requiring an additional teacher model.

\noindent\textbf{On-Policy Self-Distillation.}
On-policy self-distillation (OPSD) uses the same policy for both roles, giving
the teacher privileged information \(c\), such as hints, feedback, verifier
traces, or reference-side context \citep{zhao2026selfdistilled,penaloza2026privileged}.
The student \(\pi_\theta(a_t\mid s_t)\) and privileged teacher
\(\pi_\theta(a_t\mid s_t,c)\) share parameters \(\theta\). Each action sampled
from \(\pi_{\bar\theta}\) is scored under both contexts. The OPSD surrogate is
\begin{equation}
\mathcal{J}_{\mathrm{OPSD}}(\theta;\bar\theta)
=
-
\mathbb{E}_{\tau\sim\pi_{\bar\theta}}
\left[
\sum_t
\mathrm{KL}
\left(
\pi_\theta(\cdot\mid s_t)
\;\middle\|\;
\operatorname{sg}
\left[
\pi_{\bar\theta}(\cdot\mid s_t,c)
\right]
\right)
\right],
\end{equation}
where \(\operatorname{sg}\) denotes stop-gradient. Its local gradient at
\(\theta=\bar\theta\) is
\begin{equation}
g_{\mathrm{OPSD}}(\bar\theta)
=
\mathbb{E}_{\tau\sim\pi_{\bar\theta}}
\left[
\sum_t
\left(
\log\pi_{\bar\theta}(a_t\mid s_t,c)
-
\log\pi_{\bar\theta}(a_t\mid s_t)
\right)
\nabla_{\bar\theta}
\log\pi_{\bar\theta}(a_t\mid s_t)
\right].
\end{equation}
OPSD thus retains dense supervision without requiring an additional teacher model.

\vspace{-3pt}
\subsection{A Reward-Consistent Teacher}
\vspace{-2pt}

Conditioning the policy on privileged information makes the teacher better
informed, but does not guarantee that the resulting distillation signal improves
the student. The privileged teacher may exploit shortcuts that are unavailable
without the privileged information \(c\), assign high probability to
trajectories that are unlikely under the student, or provide unreliable guidance
after errors in a student-generated prefix. Indeed, privileged-context
distillation can degrade performance even when the teacher is more capable
\citep{chakraborty2026pedagogicalrl,kaur2026rethinking}.

These failure modes motivate characterizing a teacher by the student update it
induces, rather than by its expected reward under privileged information. For a
fixed student policy \(\pi_{\bar\theta}\), we call a teacher \(\pi_T^\star\)
reward-consistent if the local OPD update it induces matches, up to positive
scaling, the policy gradient of the student's expected return. That is, for
some \(\beta_T>0\),
\begin{equation}
g_{\mathrm{OPD}}(\bar\theta;\pi_T^\star)
=
\frac{1}{\beta_T}
\left.\nabla_\theta
\mathcal{J}_{\mathrm{RL}}(\theta)\right|_{\theta=\bar\theta}.
\end{equation}
Substituting the two gradients yields the necessary and sufficient condition
\begin{equation}
\mathbb{E}_{\tau\sim\pi_{\bar\theta}}
\left[
\sum_t
\left(
\log\frac{\pi_T^\star(a_t\mid s_t,c)}{\pi_{\bar\theta}(a_t\mid s_t)}
-
\frac{1}{\beta_T}A^{\pi_{\bar\theta}}(s_t,a_t)
\right)
\nabla_{\bar\theta}
\log\pi_{\bar\theta}(a_t\mid s_t)
\right]
=
0.
\end{equation}
A sufficient pointwise condition for this proportionality is:
\begin{equation}
\label{eq:pointwise-reward-consistency}
\log\pi_T^\star(a\mid s,c)
-
\log\pi_{\bar\theta}(a\mid s)
=
\frac{1}{\beta_T}A^{\pi_{\bar\theta}}(s,a)
-
\log Z(s),
\end{equation}
for every state-action pair in the support of the student, where the normalizing
constant is
\begin{equation}
Z(s)
=
\sum_{a'}
\pi_{\bar\theta}(a'\mid s)
\exp
\left(
\frac{1}{\beta_T}A^{\pi_{\bar\theta}}(s,a')
\right).
\end{equation}
Note that \(\log Z(s)\) does not affect the policy gradient update, as
\begin{equation}
\mathbb{E}_{a\sim\pi_{\bar\theta}(\cdot\mid s)}
\left[
\log Z(s)
\nabla_{\bar\theta}\log\pi_{\bar\theta}(a\mid s)
\right]
=
0.
\end{equation}
Exponentiating equation~(\ref{eq:pointwise-reward-consistency}) gives
\begin{equation}
\pi_T^\star(a\mid s,c)
=
\frac{1}{Z(s)}
\pi_{\bar\theta}(a\mid s)
\exp\left(
\frac{1}{\beta_T}A^{\pi_{\bar\theta}}(s,a)
\right).
\end{equation}
This exponential update from KL-regularized policy improvement
\citep{peters2010relative,abdolmaleki2018maximum} provides a sufficient
teacher for local OPD gradient proportionality.

\vspace{-3pt}
\subsection{Joint On-Policy Learning and Teaching (JOLT)}
\vspace{-2pt}

The preceding derivation characterizes a pointwise reward-consistent teacher for
a fixed student. This suggests that the privileged teacher should be trained
toward this target distribution by minimizing
\begin{equation}
\min_{\theta}
\quad
\mathrm{KL}
\left(
\pi_\theta(\cdot\mid s,c)
\;\middle\|\;
\pi_T^\star(\cdot\mid s,c)
\right).
\end{equation}
During this update, the student policy, its advantage, and the target
distribution are fixed at \(\bar\theta\). Substituting
the expression for \(\pi_T^\star\) gives
\begin{align}
&\mathrm{KL}
\left(
\pi_\theta(\cdot\mid s,c)
\;\middle\|\;
\pi_T^\star(\cdot\mid s,c)
\right)
\nonumber\\
&\quad=
\mathbb{E}_{a\sim\pi_\theta(\cdot\mid s,c)}
\left[
\log
\frac{
\pi_\theta(a\mid s,c)
}{
\operatorname{sg}\!\left[\pi_{\bar\theta}(a\mid s)\right]
}
-
\frac{1}{\beta_T}A^{\pi_{\bar\theta}}(s,a)
+
\log Z(s)
\right].
\end{align}
Since \(\log Z(s)\) is constant during the teacher update,
minimizing this divergence is equivalent to maximizing
\begin{equation}
\mathbb{E}_{a\sim\pi_\theta(\cdot\mid s,c)}
\left[
A^{\pi_{\bar\theta}}(s,a)
\right]
-
\beta_T
\mathrm{KL}
\left(
\pi_\theta(\cdot\mid s,c)
\;\middle\|\;
\operatorname{sg}\!\left[\pi_{\bar\theta}(\cdot\mid s)\right]
\right).
\end{equation}

The objective above requires student continuation advantages, which are generally
unavailable. Motivated by this construction, JOLT uses returns from the
teacher's own trajectories together with a separate token-level KL loss at
sampled teacher prefixes, yielding the following practical training surrogate:
\begin{equation}
\label{eq:teacher-rollout-objective}
\begin{aligned}
\mathcal{J}_{T}(\theta;\bar\theta)
&=
\mathbb{E}_{\tau\sim\pi_{\bar\theta}(\cdot\mid c)}
\Big[\sum\nolimits_t\big(
\operatorname{sg}[\widehat A(\tau)]
\log\pi_\theta(a_t\mid s_t,c)
\\
&\hspace{35mm}
-\beta_T\mathrm{KL}\!\left(
\pi_\theta(\cdot\mid s_t,c)
\;\middle\|\;
\operatorname{sg}[\pi_{\bar\theta}(\cdot\mid s_t)]
\right)\big)\Big].
\end{aligned}
\end{equation}
Here, \(\widehat A(\tau)\) is the within-role centered return defined below.
Sampled teacher prefixes and the student reference are held fixed during
differentiation. The gradient-proportionality result characterizes the
local student distillation update under the ideal teacher for a fixed
policy. Joint training applies teacher updates to the shared
parameters, so the combined update falls outside this guarantee.

JOLT combines this teacher surrogate
with student distillation, updating shared parameters \(\theta\). This provides
task augmentation: the model learns from original tasks alongside assisted
versions made easier by privileged information. The formulation is agnostic
to how \(c\) is constructed and accommodates existing privileged contexts.

To reduce variance in the reward-based updates, we center returns within
each role's group of \(G\) sibling trajectories sampled under the same
context \citep{shao2024deepseekmath}, leaving the separate token-level KL
loss unchanged:
\begin{equation}
\widehat A(\tau)
=
R(\tau)
-
\frac{1}{G}
\sum_{j=1}^{G}
R(\tau_j).
\end{equation}
At the rollout parameters, the local gradients admit a sampled score-function
representation. Writing
\(\Delta_t^{\bar\theta}=\log\pi_{\bar\theta}(a_t\mid s_t,c)
-\log\pi_{\bar\theta}(a_t\mid s_t)\)
for the teacher--student log-probability ratio evaluated at these parameters,
the corresponding token weights are
\begin{equation}
\widehat A_t^T
=
\operatorname{sg}\!\left[
\widehat A(\tau)-\beta_T\Delta_t^{\bar\theta}
\right],
\qquad
\widehat A_t^S
=
\operatorname{sg}\!\left[
\beta_S\Delta_t^{\bar\theta}
\right].
\end{equation}
The current-token log-ratio reproduces the gradient of the teacher's separate
KL loss in expectation at fixed prefixes and rollout parameters.
Here, \(\beta_T\) scales teacher KL and \(\beta_S\) scales student distillation.
JOLT+ adds the student's return advantage,
\(\widehat A_t^{S+}=\operatorname{sg}[\widehat A(\tau)+\beta_S\Delta_t^{\bar\theta}]\),
while leaving the teacher update unchanged.

Combining the teacher and student losses gives the shared-parameter surrogate
\begin{equation}
\mathcal{J}_{\mathrm{JOLT}}(\theta;\bar\theta)
=
\mathcal{J}_{T}(\theta;\bar\theta)
+
\beta_S\mathcal{J}_{\mathrm{OPSD}}(\theta;\bar\theta).
\end{equation}
JOLT+ adds the analogous reward-based student surrogate, combining dense
teacher supervision with direct outcome feedback on student trajectories.
In both variants, each role learns from trajectories generated under its
own context, and both roles update the shared policy.

\FloatBarrier

\begin{figure}[!t]
  \vspace{-4pt}
  \centering
  \includegraphics[width=0.87\linewidth]{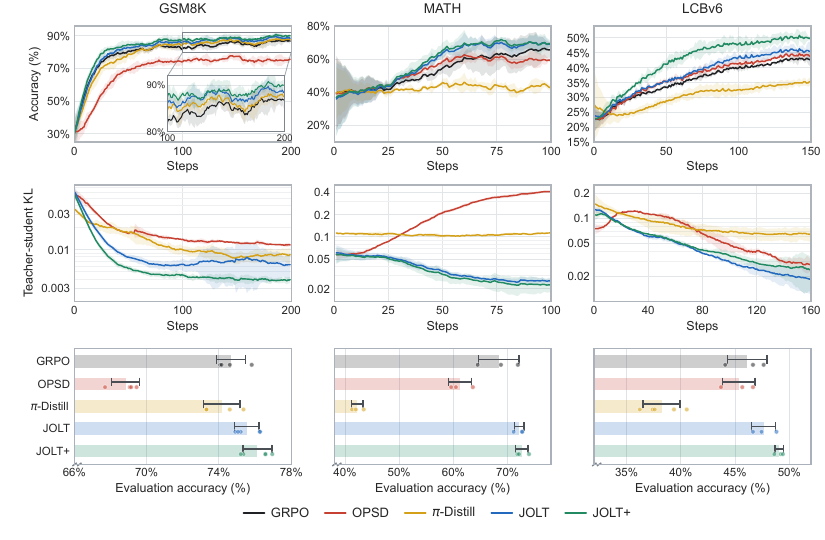}
  \par\vspace{-16pt}
  \caption{Training accuracy (top), teacher--student KL (middle), and final
  evaluation (bottom) on GSM8K, MATH, and LCBv6. JOLT improves mean evaluation
  accuracy over the distillation baselines with closer teacher--student
  agreement. Bottom-row dots show runs, and error bars show standard deviations.
  Evaluation protocols appear in Appendix~\ref{app:comparison-definitions}.}
  \label{fig:accuracy-summary-kl-grid-preview}
  \vspace{-4pt}
\end{figure}

\vspace{-3pt}
\section{Related Work}
\vspace{-3pt}

\noindent\textbf{On-Policy and Privileged Self-Distillation.}
Knowledge distillation traditionally transfers the output distribution of a
fixed teacher into a student model \citep{hinton2015distilling}. For
autoregressive models, sequence-level distillation commonly trains the student
on teacher-generated trajectories \citep{kim2016sequence}. Supervision is
therefore concentrated on prefixes generated by the teacher, which may differ
from those encountered during the student's own generation. On-policy
distillation addresses this mismatch by querying the teacher along
student-generated trajectories \mbox{\citep{agarwal2024onpolicy,song2026survey}}. Its
reverse-KL formulation provides dense supervision through the teacher-student
log-probability ratio.

Teacher supervision can also be obtained from the student model itself by
conditioning it on additional information. This builds on context distillation,
which transfers behavior elicited by additional context into a policy that acts
without that context \citep{snell2022learning}. Privileged self-distillation
combines this idea with on-policy sampling, using verified solutions,
demonstrations, textual feedback, system prompts, or prior experience to
condition the teacher
\citep{zhao2026selfdistilled,shenfeld2026selfdistillation,
hubotter2026reinforcement,ye2026onpolicycontext}. The teacher shares the
student's parameters or uses a frozen or moving-average copy, allowing this
additional information to provide dense supervision without a separately
trained teacher model.

\noindent\textbf{Reliability of Teacher Supervision.}
Recent studies have identified failures arising from teacher-student support
mismatch, high-entropy teacher predictions, privileged-context leakage, and
differences in
reasoning style
\citep{fu2026revisiting,jin2026entropy,kim2026why,
kaur2026rethinking}. These studies show that teacher guidance can be unreliable
on student-generated prefixes, even when the teacher performs well on its own
trajectories.

To assess whether teacher supervision helps the student,
\citet{armandpour2026unmasking} compare distillation gradients with estimated
success-improving gradients. \citet{agrawal2026distil} examine this connection
through policy-improvement analysis, deriving guarantees for a distributional
DAgger objective under stated assumptions. Other approaches modify how reward
and dense supervision are combined. RLSD uses outcome rewards to determine
update directions while allowing teacher-student disagreement to modulate their
token-level magnitudes \citep{yang2026selfdistilledrlvr}. G-OPD instead
interprets the teacher-reference log-probability ratio as an implicit dense
reward, allowing its strength to be adjusted independently of KL regularization
\citep{yang2026learningbeyond}. These approaches assess or modify the student
update induced by a teacher. We instead characterize teacher distributions
under which the OPD update follows the student's reward policy gradient, using
this condition to motivate teacher training.

\noindent\textbf{Learned Teachers and Policy Improvement.}
Teacher training can improve supervision by accounting for the student's
learning needs.
Reinforcement-Learned Teachers optimize explanations using the resulting
student performance \citep{cetin2025reinforcement}, while SOAR rewards problem
generation that improves the student on difficult tasks
\mbox{\citep{sundaram2026teaching}}. Pedagogical RL trains a privileged teacher using
task success together with a spike-sensitive surprise penalty that measures
learnability under the student \citep{chakraborty2026pedagogicalrl}. Teacher
optimization can also be integrated with student learning. \(\pi\)-Distill
jointly trains privileged teacher and student roles within a shared model,
using a KL-regularized teacher objective. However, its student update remains
off-policy, relying on teacher-generated trajectories rather than the student's
own generations \citep{penaloza2026privileged}. The exponential reweighting in
our teacher construction connects it to KL-regularized policy improvement and
related self-distillation formulations
\citep{peters2010relative,abdolmaleki2018maximum,peng2019advantage,
geist2019theory,ding2026hdpo,yu2026pbsd}.

\begingroup
\setlength{\parskip}{4pt}
\raggedbottom
\Needspace{4\baselineskip}
\vspace{-3pt}
\section{Experiments}
\vspace{-3pt}
\label{sec:experiments}

\begin{wraptable}{R}{0.55\textwidth}
  \vspace{-10pt}
  \centering\small
  \setlength{\abovecaptionskip}{0pt}
  \setlength{\belowcaptionskip}{4pt}
  \caption{Evaluation (\%): MATH/LCBv6 accuracy and AppWorld
  development-set state-test pass rate. Mean $\pm$ standard deviation
  across runs. Run counts and protocols are in
  Appendix~\ref{app:experimental-details}.}
  \label{tab:evaluation-summary}
  \setlength{\tabcolsep}{3pt}
  \renewcommand{\arraystretch}{1.15}
  \begin{tabular}{@{}lccc@{}}
    \toprule
    \textbf{Method} & \textbf{MATH} & \textbf{LCBv6} & \textbf{AppWorld} \\
    \midrule
    GRPO & $68.36 \pm 3.73$ & $46.13 \pm 1.83$ & $62.20 \pm 2.83$ \\
    \midrule
    OPSD & $61.20 \pm 2.10$ & $45.34 \pm 1.50$ & $29.49 \pm 2.48$ \\
    $\pi$-Distill & $42.25 \pm 1.10$ & $38.23 \pm 1.72$ & $59.85 \pm 2.97$ \\
    \midrule
    JOLT & $72.14 \pm 0.90$ & $47.64 \pm 1.10$ & $51.92 \pm 5.30$ \\
    \textbf{JOLT+} & $\mathbf{72.59 \pm 1.17}$ & $\mathbf{49.08 \pm 0.41}$ & $\mathbf{72.09 \pm 6.01}$ \\
    \bottomrule
  \end{tabular}
  \vspace{-8pt}
\end{wraptable}

We evaluate joint teacher--student training both without direct student
reward and alongside reward-based student learning. Our experiments span mathematical reasoning,
code generation, tool use, and long-horizon terminal tasks. Across these
settings, the teacher receives different forms of privileged information,
including verified answers, worked solutions, execution feedback, and
summaries of previous attempts. We compare against
GRPO~\citep{shao2024deepseekmath}, OPSD~\citep{zhao2026selfdistilled}, and
$\pi$-Distill~\citep{penaloza2026privileged}, tuning hyperparameters for each
method. Tables~\ref{tab:evaluation-summary} and~\ref{tab:terminal-evaluation}
summarize evaluation performance, while Appendix~\ref{app:experimental-details}
provides experimental configurations, prompts, and tuning details.

Table~\ref{tab:evaluation-summary} reports means and standard deviations
across runs. Appendix~\ref{app:experimental-details} gives run counts and other
uncertainty conventions. We apply exponential moving-average smoothing
only to training curves.

\begin{figure}[!t]
  \vspace{-4pt}
  \centering
  \includegraphics[width=0.75\linewidth]{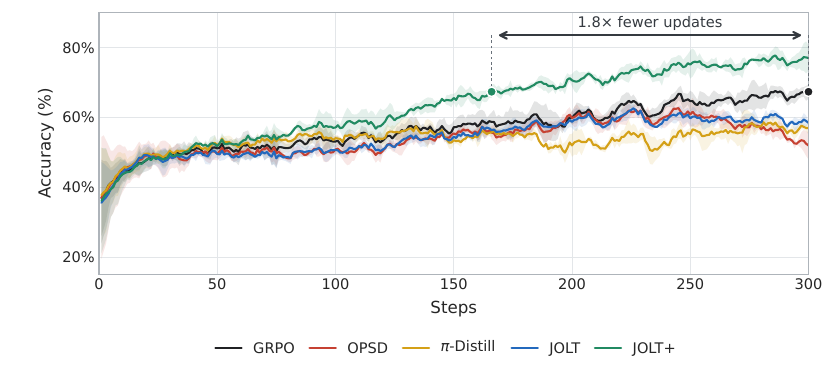}
  \par\vspace{-16pt}
  \caption{Training performance on AppWorld. JOLT+ reaches GRPO's final
  training performance using approximately $1.8\times$ fewer updates, then
  continues improving. Combining teacher supervision with
  student outcome rewards outperforms both distillation baselines.}
  \label{fig:appworld-training}
  \vspace{-4pt}
\end{figure}

\noindent\textbf{GSM8K.}
We train DeepSeek-V2-Lite~\citep{deepseekai2024deepseekv2} on
GSM8K~\citep{cobbe2021training}, supplying the teacher with the verified
answer to each problem. As shown in
Figure~\ref{fig:accuracy-summary-kl-grid-preview} (left), OPSD improves early
in training but plateaus, whereas JOLT and JOLT+ continue improving as
teacher--student disagreement decreases. Final evaluations in the bottom-left
panel place JOLT close to tuned GRPO without direct student reward, with
JOLT+ highest in mean accuracy.

\noindent\textbf{MATH.}
We train Qwen3-0.6B~\citep{qwen2025qwen3} on
MATH~\citep{hendrycks2021math}, supplying worked reference solutions to the
teacher following \citet{zhao2026selfdistilled}.
Figure~\ref{fig:accuracy-summary-kl-grid-preview} (middle) shows that JOLT and
JOLT+ improve more rapidly than GRPO and reach higher training accuracy,
while OPSD plateaus earlier. Teacher--student disagreement grows under OPSD but
decreases under joint training. The bottom row shows both JOLT variants
above OPSD and $\pi$-Distill in mean held-out accuracy.
Table~\ref{tab:evaluation-summary} also places JOLT above GRPO without direct
student reward, with JOLT+ highest.
Appendix~\ref{app:teacher-ablation-details} examines the contributions of the
teacher's reward and KL terms and shows that JOLT+ outperforms OPSD augmented
with student outcome rewards.

\begin{figure}[!t]
  \vspace{-6pt}
  \centering
  \includegraphics[width=0.96\linewidth]{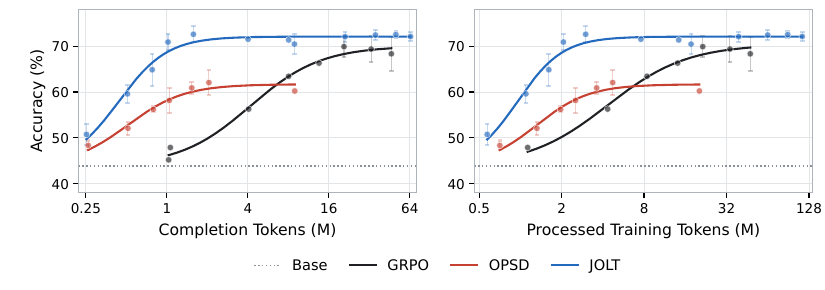}
  \par\vspace{-10pt}
  \caption{Performance scaling with training token budget, as measured in
  completion tokens (left) and processed training tokens (right). On MATH,
  JOLT reaches high accuracy with fewer tokens than GRPO while surpassing
  OPSD's early plateau under both token counters.
  Appendix~\ref{app:token-details} details token accounting, curve fitting,
  and the budget comparison.}
  \label{fig:math-token-efficiency}
  \vspace{-4pt}
\end{figure}

\begin{figure}[!b]
  \vspace{-4pt}
  \centering
  \includegraphics[width=\linewidth]{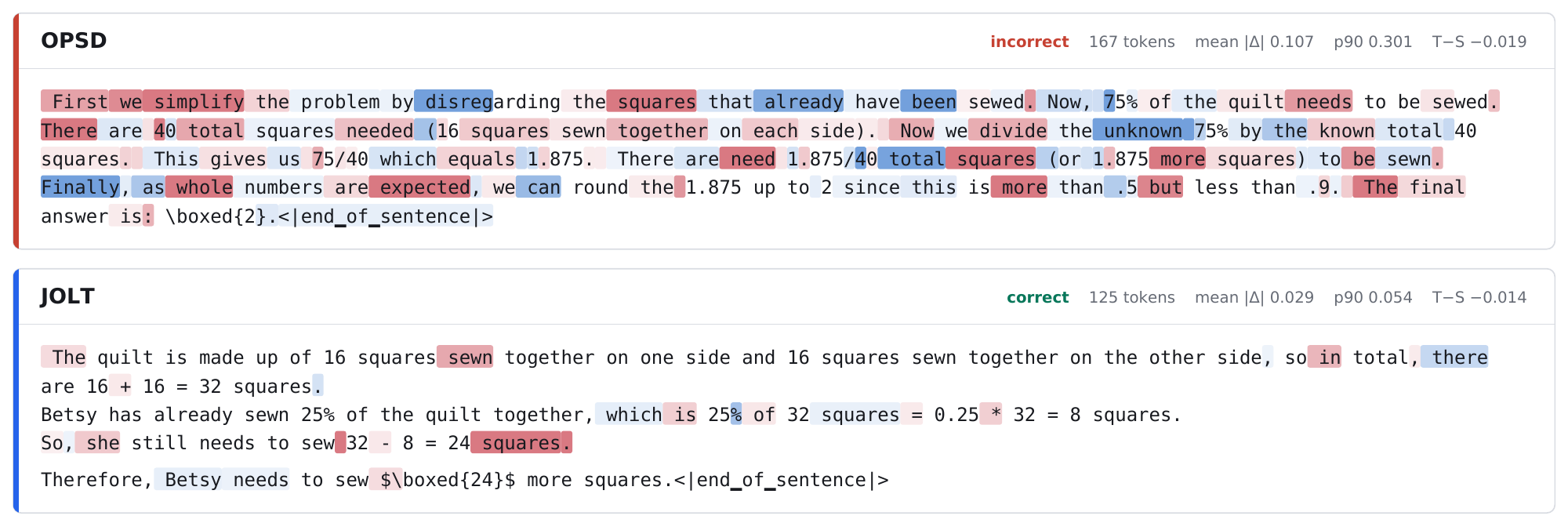}
  \par\vspace{-8pt}
  \caption{Teacher--student log-probability differences along responses to
  the same GSM8K problem under OPSD (top) and JOLT (bottom). Red indicates higher teacher log probability,
  while blue indicates lower teacher log probability. In this example,
  OPSD exhibits disagreement
  throughout much of its response, while JOLT shows smaller differences
  concentrated at fewer token positions. Appendix~\ref{app:token-alignment}
  extends the comparison to $\pi$-Distill and JOLT+.}
  \label{fig:token-logprob-example}
\end{figure}

\noindent\textbf{LiveCodeBench-V6 (LCBv6).}
We next examine whether JOLT can use privileged information constructed
during training. Following SDPO's coding protocol~\citep{hubotter2026reinforcement},
we train Qwen3-8B~\citep{qwen2025qwen3} on 131 LCBv6
problems~\citep{jain2024livecodebench} and evaluate on the complete test
suites for those problems, including test cases withheld during training.
We construct teacher guidance from the model's
own attempts: the teacher receives a successful sibling solution
when available and execution feedback otherwise. As shown in
Figure~\ref{fig:accuracy-summary-kl-grid-preview} (right), JOLT improves over
the distillation baselines during training, with further gains from student
outcome rewards in JOLT+. The bottom row shows higher mean
accuracy for JOLT than GRPO and both distillation baselines, with JOLT+
highest. These results extend joint training to settings where privileged
guidance comes from sampled attempts and their outcomes.

\noindent\textbf{AppWorld.}
AppWorld~\citep{trivedi2024appworld} tests tool use through interactions with
stateful applications. We train Qwen3-4B-Instruct~\citep{qwen2025qwen3},
constructing privileged guidance from summaries of previous attempts
following \citet{wang2026skillsd}. These summaries record successful
strategies, mistakes, and suggested workflows. Following Skill-SD, we
report state-test pass rate on the 57-task development split, averaging
the fraction of tests passed per task.
Figure~\ref{fig:appworld-training} shows that OPSD deteriorates later in
training, while JOLT remains competitive with $\pi$-Distill but below GRPO.
Combining teacher supervision with student outcome rewards produces stronger
results: JOLT+ reaches GRPO's final training reward using approximately
$1.8\times$ fewer training updates and continues improving.
As reported in Table~\ref{tab:evaluation-summary}, JOLT's mean development-set
state-test pass rate remains below $\pi$-Distill and GRPO, whereas JOLT+
achieves the highest mean.

\begin{wraptable}{R}{0.55\textwidth}
  \vspace{0pt}
  \centering\small
  \setlength{\abovecaptionskip}{0pt}
  \setlength{\belowcaptionskip}{4pt}
  \caption{Terminal-Bench 2.1 pass@$k$ (\%) after TMax training,
  using one training run per method.}
  \label{tab:terminal-evaluation}
  \setlength{\tabcolsep}{4pt}
  \renewcommand{\arraystretch}{1.15}
  \begin{tabular}{@{}lrrrr@{}}
    \toprule
    \textbf{Method} & \textbf{pass@1} & \textbf{pass@4} & \textbf{pass@16} & \textbf{pass@32} \\
    \midrule
    Base & 5.18 & 10.84 & 18.35 & 22.81 \\
    GRPO & 8.25 & 17.04 & 26.91 & 30.34 \\
    \midrule
    OPSD & 2.82 & 6.96 & 13.59 & 16.90 \\
    $\pi$-Distill & 4.99 & 10.50 & 18.30 & 22.31 \\
    \midrule
    JOLT & 6.56 & 13.56 & 22.06 & 25.69 \\
    \textbf{JOLT+} & \textbf{10.96} & \textbf{20.12} & \textbf{29.99} & \textbf{34.83} \\
    \bottomrule
  \end{tabular}
  \vspace{-8pt}
\end{wraptable}

\noindent\textbf{Terminal-Bench.}
We train Qwen3-30B-A3B-Instruct~\citep{qwen2025qwen3} on
TMax~\citep{ivison2026tmax} and evaluate on
Terminal-Bench 2.1~\citep{merrill2026terminalbench} using TMax's released
agent and evaluation settings. During training, the teacher receives
private verifier information in stateful shell environments. Evaluation uses the
unprivileged student. Table~\ref{tab:terminal-evaluation} reports
pass@$k$~\citep{chen2021evaluating} for $k\in\{1,4,16,32\}$. JOLT improves over
the base model and both distillation baselines at every reported $k$,
although GRPO achieves higher scores.
JOLT+ achieves the highest pass@$k$ across all four reported values,
suggesting that teacher supervision can complement student outcome
rewards on long-horizon tasks.

\begin{figure}[!t]
  \vspace{-4pt}
  \centering
  \includegraphics[width=0.91\linewidth]{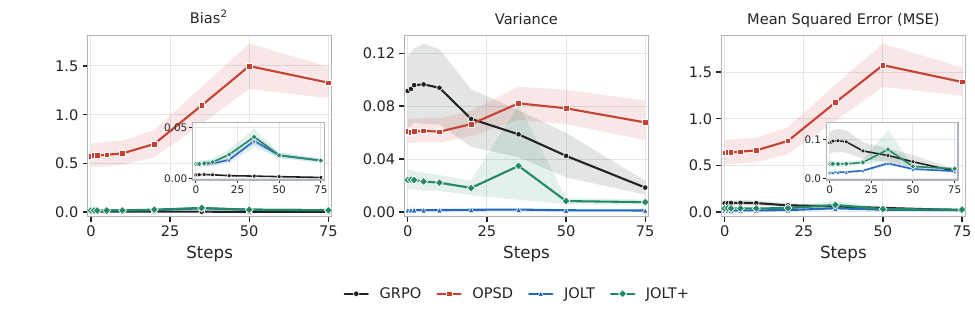}
  \par\vspace{-10pt}
  \caption{Empirical squared bias (left), variance (middle), and mean squared
  error (right) of weighted student gradients along each method's own
  training trajectory. Bias and MSE use Monte Carlo reward-gradient references.
  GRPO's error is dominated by variance,
  whereas OPSD develops increasing empirical bias. JOLT and JOLT+ retain low
  variance with smaller empirical bias than OPSD.
  Bands show pointwise 95\% batch-bootstrap intervals.
  See Appendix~\ref{app:gradient-details} for details.}
  \label{fig:gradient-student-overlay}
  \vspace{-10pt}
\end{figure}

\noindent\textbf{Token Efficiency and Scaling.}
Beyond final accuracy, we are interested in how performance scales with
training tokens, as a proxy for training compute. To this end, we evaluate
models trained under different training token budgets. Following
\citet{khatri2025art}, we fit a sigmoidal relationship between accuracy and
log training-token budget. As shown in
Figure~\ref{fig:math-token-efficiency}, OPSD initially learns more
efficiently than GRPO, but its advantage disappears as its performance
plateaus while GRPO continues improving. Meanwhile, JOLT exhibits early
sample efficiency while reaching substantially higher accuracy compared to
GRPO. This pattern holds for both completion tokens and processed training
tokens, showing that joint training improves performance across training
budgets. Appendix~\ref{app:token-details} details token accounting, curve
fitting, and savings across accuracy targets.

\noindent\textbf{Student--Teacher Alignment.}
Beyond the aggregate student--teacher KL measurements shown in
Figure~\ref{fig:accuracy-summary-kl-grid-preview}, we examine how the teacher
and student disagree along individual generations. In the example shown in
Figure~\ref{fig:token-logprob-example}, the privileged teacher under OPSD
assigns markedly different probabilities to tokens throughout much of the
student's response. Meanwhile, under JOLT, these differences are smaller
and concentrated at fewer locations. These observations illustrate
differences in both the overall magnitude of contextual disagreement and
its distribution across the tokens receiving supervision.
Appendix~\ref{app:token-alignment} extends this token-level comparison to
$\pi$-Distill and JOLT+.

\noindent\textbf{Gradient Bias--Variance Analysis.}
To better understand the student learning signals encountered during training,
we examine their empirical bias and variance at checkpoints along each
method's own trajectory. We estimate bias relative to a Monte Carlo
reward-gradient reference obtained by averaging independent REINFORCE
gradient estimates~\citep{williams1992simple}, and variance across repeated
student-gradient estimates. As shown in
Figure~\ref{fig:gradient-student-overlay}, GRPO's measured error is dominated
by variance, whereas OPSD exhibits low variance but increasing empirical bias
during training. JOLT and JOLT+ retain low variance while showing substantially
smaller empirical bias than OPSD. These measurements depend on gradient
scale and learned policies. They do not isolate teacher training's effect
or establish reward consistency of the complete joint update.
Appendix~\ref{app:gradient-details} gives definitions and a comparison of
normalized gradient directions.
\par
\endgroup

\vspace{-3pt}
\section{Conclusion}
\vspace{-3pt}

We introduced JOLT, a method that jointly trains a single policy as a privileged
teacher and an unprivileged student. Our analysis connects the teacher's
supervision to the student's reward objective, motivating a teacher objective that
balances task performance with proximity to the current student. Both roles
learn from their own trajectories, and the formulation accommodates different
sources of privileged information. Experiments demonstrate improvements in both
the effectiveness and efficiency of self-distillation: joint training reaches
competitive performance without direct student reward and achieves high accuracy
with fewer training tokens. Incorporating student rewards further strengthens
this approach by combining dense teacher supervision with direct feedback on the
student's own generations. The token-level visualizations show closer
teacher--student agreement under JOLT, while the gradient analysis characterizes
the variability and empirical reward-gradient discrepancies of the student
updates along training.
More broadly, this provides a way to
supplement sparse rewards with dense supervision that adapts to the model's
current capabilities, without requiring a separate teacher.

\FloatBarrier
\endgroup

\subsection*{Acknowledgements}

We thank Andrew Gordon Wilson, Ziqing Hu, Hang Wu, Senzeyu Zhang, Dongqi Wu,
Yuan Liang, Yunfan Zhong, Lequn Chen, and Denis Yarats for insightful
discussions that helped shape this work. We are also grateful to Ziqing Hu,
Yunfan Zhong, Zhihao Wang, and Lequn Chen for their support with the training
infrastructure that enabled our experiments.

\subsection*{AI Use}

The research ideas originated with the authors. Generative AI tools assisted
with implementing and debugging methods and baselines, analysis and
figure-generation code, literature search and synthesis, manuscript drafting
and editing, and LaTeX preparation. They also supported discussion and critique
of derivations, experimental design, and result interpretation. The authors
thoroughly reviewed and iteratively refined the implementations, and
cross-checked quantitative summaries and reproduced figures against experiment
records. AI was not used to generate or clean benchmark datasets. The authors
are responsible for the final text, mathematical claims, code, and reported
results.

\clearpage
\appendix
\raggedbottom
\setlength{\textfloatsep}{12pt plus 2pt minus 2pt}
\setlength{\intextsep}{12pt plus 2pt minus 2pt}
\setlength{\floatsep}{10pt plus 2pt minus 2pt}
\section{Experimental Details and Additional Results}
\label{app:experimental-details}
\label{app:additional-results}
\label{app:comparison-definitions}

\begin{figure}[!b]
  \centering
  \includegraphics[width=\linewidth]{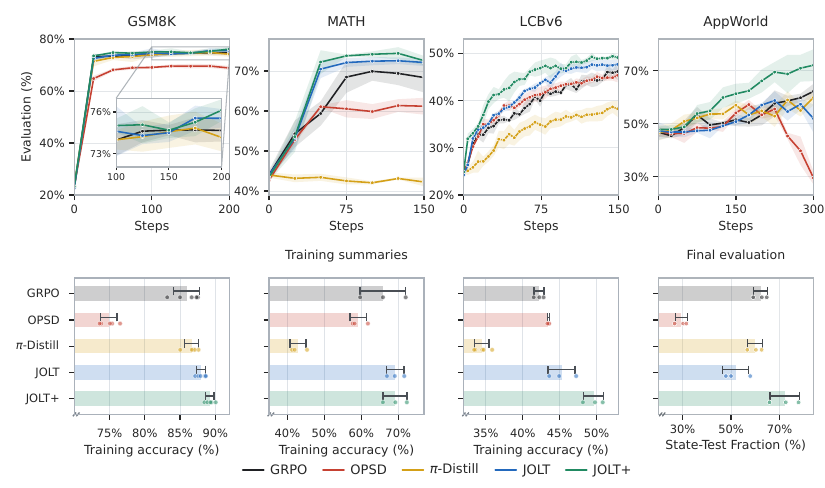}
  \par\vspace{-4pt}
  \caption{Evaluation curves on GSM8K, MATH, LCBv6, and AppWorld (top), with
  training-accuracy summaries for the first three tasks and final AppWorld
  evaluation below. Final evaluation summaries for the first three tasks
  appear in Figure~\ref{fig:accuracy-summary-kl-grid-preview}.
  LCBv6 evaluates training problems on complete test suites; AppWorld reports
  development-set state-test pass rate. Bands and error bars show standard
  deviations across runs; dots show individual runs.}
  \label{fig:evaluation-summary-grid}
  \vspace{-6pt}
\end{figure}

This appendix provides experimental settings, prompts, and additional
results. Table~\ref{tab:hyperparameter-summary} summarizes training
hyperparameters, and Figure~\ref{fig:evaluation-summary-grid} shows evaluation
curves and training summaries across benchmarks. Table~\ref{tab:evaluation-summary} reports
means and sample standard deviations across independent training
runs, using three runs per configuration, except
for $\pi$-Distill on LCBv6, which uses five.
The evaluation panels in Figures~\ref{fig:accuracy-summary-kl-grid-preview}
and~\ref{fig:evaluation-summary-grid} use these same runs for MATH,
LCBv6, and AppWorld. GSM8K evaluation uses four runs for GRPO, OPSD, and
$\pi$-Distill, and five for JOLT and JOLT+. Evaluation bands and error bars
show standard deviations across runs.
The training summaries in Figure~\ref{fig:evaluation-summary-grid} also
show standard deviations, using five runs for GSM8K and three for MATH
and LCBv6, except for $\pi$-Distill on LCBv6, which uses five.
Other experiments use five independent training seeds and report
normal-approximation 95\% confidence intervals (mean $\pm 1.96$ standard
errors), except for Terminal-Bench and the gradient analysis, which use
one run per method.
The gradient analysis uses batch-bootstrap intervals, as described in
Appendix~\ref{app:gradient-details}. Exponential moving-average smoothing is
applied only to training curves. Appendix~\ref{app:comparison-uncertainty}
assesses consistency across settings using the average probability of
improvement and stratified bootstrap intervals.

\paragraph{Hyperparameter and checkpoint selection.}
We select hyperparameters separately for each method using
training curves and final training performance. The selected
configurations are fixed before the repeated-seed comparisons.
The main result tables and final-checkpoint summaries report
the final scheduled checkpoint of each training run.
Figure~\ref{fig:accuracy-summary-kl-grid-preview} shows final evaluations
at steps 200, 150, and 150 for GSM8K, MATH, and LCBv6, respectively; its MATH
training and KL curves display only the first 100 steps.

\begin{table}[H]
  \centering
  \caption{Selected training hyperparameters. $B$ denotes the number of
  problems per update and $G$ the group size per active role.
  On GSM8K and MATH, JOLT/JOLT+ use four teacher and four student rollouts
  per problem, while GRPO and OPSD use eight student rollouts. Learning
  rates are initial values where a schedule is used.}
  \label{tab:hyperparameter-summary}
  \begingroup
  \small
  \setlength{\tabcolsep}{4pt}
  \renewcommand{\arraystretch}{1.06}
  \begin{tabular}{@{}llccccc@{}}
    \toprule
    Method & Hyperparameter & GSM8K & MATH & LCBv6 & AppWorld & TMax \\
    \midrule
    \multirow{3}{*}{GRPO}
      & Learning Rate & $1\times10^{-6}$ & $3\times10^{-6}$ & $1\times10^{-6}$ & $5\times10^{-7}$ & $1\times10^{-6}$ \\
      & Batch Size $B$ & 8 & 8 & 32 & 4 & 8 \\
      & Group Size $G$ & 8 & 8 & 8 & 4 & 32 \\
      & Num Steps & 200 & 150 & 150 & 300 & 50 \\
    \midrule
    \multirow{3}{*}{OPSD}
      & Learning Rate & $1\times10^{-7}$ & $2\times10^{-6}$ & $1\times10^{-6}$ & $1\times10^{-7}$ & $2.5\times10^{-7}$ \\
      & Batch Size $B$ & 8 & 8 & 32 & 4 & 8 \\
      & Group Size $G$ & 8 & 8 & 8 & 4 & 32 \\
      & Num Steps & 200 & 150 & 150 & 300 & 50 \\
    \midrule
    \multirow{5}{*}{$\pi$-Distill}
      & Learning Rate & $1\times10^{-6}$ & $2.5\times10^{-7}$ & $1\times10^{-6}$ & $5\times10^{-7}$ & $5\times10^{-7}$ \\
      & Batch Size $B$ & 8 & 8 & 32 & 4 & 8 \\
      & Group Size $G$ & 8 & 8 & 8 & 4 & 32 \\
      & Num Steps & 200 & 150 & 150 & 300 & 50 \\
      & Teacher Mixing $w_T$ & 0.5 & 0.5 & 0.1 & 0.5 & 0.5 \\
      & Teacher KL $\beta_T$ & 0.25 & 0 & 0.1 & 0.25 & 0.25 \\
    \midrule
    \multirow{5}{*}{JOLT/JOLT+}
      & Learning Rate & $1\times10^{-6}$ & $2\times10^{-6}$ & $1\times10^{-6}$ & $1\times10^{-7}$ & $1\times10^{-6}$ \\
      & Batch Size $B$ & 8 & 8 & 8 & 2 & 8 \\
      & Group Size $G$ & 4 & 4 & 8 & 4 & 16 \\
      & Num Steps & 200 & 150 & 150 & 300 & 50 \\
      & Teacher KL $\beta_T$ & 0.03 & 0.001 & 0.001 & 0.001 & 0.001 \\
      & Student Distillation $\beta_S$ & 0.1 & 0.1 & 0.1 & 0.1 & 0.1 \\
    \bottomrule
  \end{tabular}
  \endgroup
  \par\smallskip
\end{table}

\clearpage

\begin{figure}[!t]
  \centering
  \begin{minipage}[t]{0.50\linewidth}
    \centering
    \includegraphics[width=\linewidth]{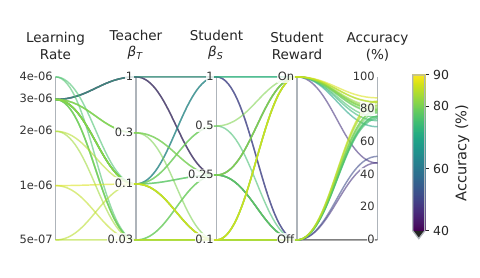}
  \end{minipage}\hfill
  \begin{minipage}[t]{0.48\linewidth}
    \centering
    \includegraphics[width=\linewidth]{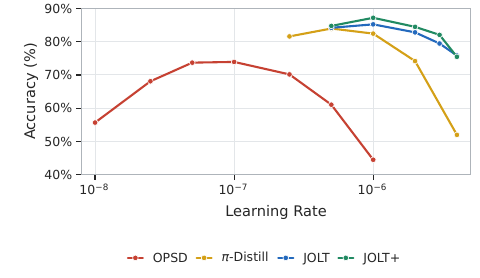}
  \end{minipage}
  \par\vspace{-4pt}
  \caption{Hyperparameter sensitivity on GSM8K, covering JOLT configurations
  (left) and learning rates across methods (right).}
  \label{fig:appendix-sensitivity}
  \vspace{-6pt}
\end{figure}

\subsection{GSM8K}
\label{app:gsm8k-details}

\paragraph{Data and evaluation.}
We use the official GSM8K training and test splits, containing
7,473 and 1,319 problems, respectively. Training samples
responses at temperature 1.0, while evaluation generates one
response per test problem using greedy decoding. We use
top-$p=1.0$ during training and evaluation, with a maximum
sequence length of 8,192 tokens and a completion limit of
4,096 tokens.

\paragraph{Sensitivity.}
Figure~\ref{fig:appendix-sensitivity} summarizes the coefficient and
method-specific learning-rate sweeps used to select the confirmatory settings.
The strongest JOLT configurations use a learning rate near \(10^{-6}\) and
coupling coefficients below one. Unit teacher and student coefficients are
unstable in this setting, and increasing the learning rate beyond the selected
range degrades all methods.

\begin{figure}[!b]
  \centering
  \begin{minipage}[t]{0.485\linewidth}
    \centering
    \includegraphics[width=\linewidth,height=0.32\textheight,keepaspectratio]{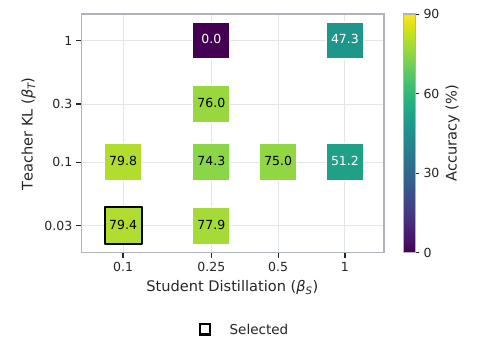}
    \par\smallskip
    {\small (a) JOLT}
  \end{minipage}\hfill
  \begin{minipage}[t]{0.485\linewidth}
    \centering
    \includegraphics[width=\linewidth,height=0.32\textheight,keepaspectratio]{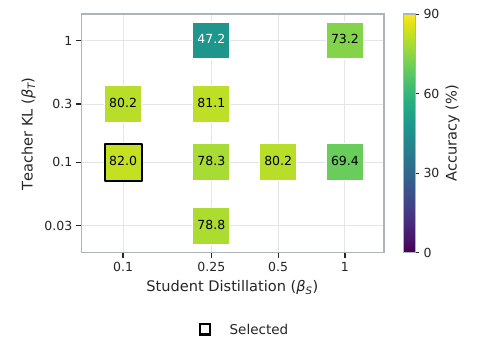}
    \par\smallskip
    {\small (b) JOLT+}
  \end{minipage}
  \caption{GSM8K accuracy across teacher--student coupling strengths for
  (a) JOLT and (b) JOLT+.}
  \label{fig:coefficient-sweeps}
\end{figure}

The exploratory coefficient sweeps in Figure~\ref{fig:coefficient-sweeps}
are distinct from the five-seed confirmatory comparison.

\begingroup
\lstdefinestyle{joltprompt}{
  language={},
  basicstyle=\fontsize{8.5}{10.2}\fontencoding{T1}\fontfamily{lmtt}\selectfont,
  columns=fullflexible,
  keepspaces=true,
  breaklines=true,
  breakatwhitespace=true,
  breakindent=0pt,
  breakautoindent=false,
  showstringspaces=false,
  showspaces=false,
  showtabs=false,
  upquote=true,
  tabsize=2,
  frame=none,
  aboveskip=0pt,
  belowskip=0pt,
  xleftmargin=7pt,
  xrightmargin=7pt,
  resetmargins=true
}
\ifdefined\joltpromptbox\else\newsavebox{\joltpromptbox}\fi
\newenvironment{joltprompt}[2][black]{%
  \par\addvspace{7pt}%
  \begingroup
  \colorlet{promptaccent}{#1}%
  \setlength{\fboxsep}{0pt}%
  \setlength{\fboxrule}{0.45pt}%
  \begin{lrbox}{\joltpromptbox}%
    \begin{minipage}{\dimexpr\linewidth-2\fboxrule\relax}
      \setlength{\fboxsep}{6pt}%
      \noindent\colorbox{promptaccent!9}{%
        \parbox{\dimexpr\linewidth-2\fboxsep\relax}{%
          \raggedright\fontsize{9}{10.5}\selectfont\bfseries
          \color{promptaccent!75!black}#2}}\par
      \nointerlineskip
      {\color{promptaccent!35}\hrule height 0.3pt}%
      \vspace{6pt}%
}{%
      \vspace{6pt}%
    \end{minipage}%
  \end{lrbox}%
  \noindent\fcolorbox{promptaccent!55}{promptaccent!2}{\usebox{\joltpromptbox}}%
  \par\endgroup\addvspace{8pt}%
}

\paragraph{Prompts and privileged context.}
\label{app:prompts-gsm8k}
Both roles receive the original question as a user message, without few-shot
examples. The teacher's system message additionally supplies the dataset's
final answer.

\begin{joltprompt}[joltstudent]{Student System Message}
\begin{lstlisting}[style=joltprompt]
You will be given a problem. Please reason step by step, and put your final answer within \boxed{}.
\end{lstlisting}
\end{joltprompt}

\begin{joltprompt}[joltteacher]{Teacher System Message}
\begin{lstlisting}[style=joltprompt]
You will be given a problem. Please reason step by step, and put your final answer within \boxed{}.

Private reference answer: {answer}. This answer has been independently verified. Derive the solution yourself, use the reference to resolve uncertainty, and make your single boxed final answer match it.
\end{lstlisting}
\end{joltprompt}
\begin{joltprompt}{User Message (Both Roles)}
\begin{lstlisting}[style=joltprompt]
{question}
\end{lstlisting}
\end{joltprompt}
\endgroup

\FloatBarrier
\subsection{MATH}
\label{app:math-details}
\suppressfloats[t]

\paragraph{Data and evaluation.}
We use the full official MATH training and test splits,
containing 7,500 and 5,000 problems, respectively. Training
and evaluation use temperature 1.0, top-$p=1.0$, and a
completion limit of 16,384 tokens. Evaluation samples one
unprivileged student response per test problem. These
evaluation settings apply to the main results, token-scaling
experiments, and teacher-objective ablations.

\begingroup
\lstdefinestyle{joltprompt}{
  language={},
  basicstyle=\fontsize{8.5}{10.2}\fontencoding{T1}\fontfamily{lmtt}\selectfont,
  columns=fullflexible,
  keepspaces=true,
  breaklines=true,
  breakatwhitespace=true,
  breakindent=0pt,
  breakautoindent=false,
  showstringspaces=false,
  showspaces=false,
  showtabs=false,
  upquote=true,
  tabsize=2,
  frame=none,
  aboveskip=0pt,
  belowskip=0pt,
  xleftmargin=7pt,
  xrightmargin=7pt,
  resetmargins=true
}
\ifdefined\joltpromptbox\else\newsavebox{\joltpromptbox}\fi
\newenvironment{joltprompt}[2][black]{%
  \par\addvspace{7pt}%
  \begingroup
  \colorlet{promptaccent}{#1}%
  \setlength{\fboxsep}{0pt}%
  \setlength{\fboxrule}{0.45pt}%
  \begin{lrbox}{\joltpromptbox}%
    \begin{minipage}{\dimexpr\linewidth-2\fboxrule\relax}
      \setlength{\fboxsep}{6pt}%
      \noindent\colorbox{promptaccent!9}{%
        \parbox{\dimexpr\linewidth-2\fboxsep\relax}{%
          \raggedright\fontsize{9}{10.5}\selectfont\bfseries
          \color{promptaccent!75!black}#2}}\par
      \nointerlineskip
      {\color{promptaccent!35}\hrule height 0.3pt}%
      \vspace{6pt}%
}{%
      \vspace{6pt}%
    \end{minipage}%
  \end{lrbox}%
  \noindent\fcolorbox{promptaccent!55}{promptaccent!2}{\usebox{\joltpromptbox}}%
  \par\endgroup\addvspace{8pt}%
}

\label{app:prompts-math}
We train Qwen3-0.6B~\citep{qwen2025qwen3} on MATH~\citep{hendrycks2021math}. We supply privileged information to the teacher by providing a worked reference solution. Both roles use the same system message and problem instruction. For the
teacher, the dataset's worked solution is appended to the user message,
after a blank line. Braced names denote substituted fields. Qwen3 thinking
is disabled; evaluation uses the student messages without the solution.

\begin{joltprompt}{System Message}
\begin{lstlisting}[style=joltprompt]
You are a helpful assistant.
\end{lstlisting}
\end{joltprompt}
\begin{joltprompt}[joltstudent]{Student User Message}
\begin{lstlisting}[style=joltprompt]
{problem}
Please reason step by step, and put your final answer within \boxed{}.
\end{lstlisting}
\end{joltprompt}
\begin{joltprompt}[joltteacher]{Teacher-Only Addition to the User Message}
\begin{lstlisting}[style=joltprompt]
Here is a verified reference solution to the problem:
=== Reference Solution Begin ===
{reference_solution}
=== Reference Solution End ===

Understand the reference solution and its key steps. Then solve the problem independently in your own words, check the reasoning, and derive the same final answer.
\end{lstlisting}
\end{joltprompt}
\endgroup

\FloatBarrier
\subsection{LiveCodeBench-V6}
\label{app:evaluation-candidates}
\suppressfloats[t]

\paragraph{Data and evaluation.}
Following the coding setup of \citet{hubotter2026reinforcement},
we use 131 LiveCodeBench v6 problems released between February
and April 2025. For each problem, we randomly select approximately
50\% of its upstream private tests as training-visible tests,
using NumPy seed 0. This produces 2,421 training-visible cases
and leaves 2,431 cases withheld from training. Evaluation follows
SDPO's released implementation and uses the complete original
suite of 4,852 test cases for these same problems, including
both the training-visible and withheld cases. We generate four
unprivileged student responses per problem and report average
accuracy across responses and problems. Evaluation uses
temperature 0.6 and top-$p=0.95$, with maximum prompt and
response lengths of 2,048 and 8,192 tokens, respectively.
This protocol evaluates
solutions to training problems against an expanded test suite
containing unseen test inputs.

\begingroup
\lstdefinestyle{joltprompt}{
  language={},
  basicstyle=\fontsize{8.5}{10.2}\fontencoding{T1}\fontfamily{lmtt}\selectfont,
  columns=fullflexible,
  keepspaces=true,
  breaklines=true,
  breakatwhitespace=true,
  breakindent=0pt,
  breakautoindent=false,
  showstringspaces=false,
  showspaces=false,
  showtabs=false,
  upquote=true,
  tabsize=2,
  frame=none,
  aboveskip=0pt,
  belowskip=0pt,
  xleftmargin=7pt,
  xrightmargin=7pt,
  resetmargins=true
}
\ifdefined\joltpromptbox\else\newsavebox{\joltpromptbox}\fi
\newenvironment{joltprompt}[2][black]{%
  \par\addvspace{7pt}%
  \begingroup
  \colorlet{promptaccent}{#1}%
  \setlength{\fboxsep}{0pt}%
  \setlength{\fboxrule}{0.45pt}%
  \begin{lrbox}{\joltpromptbox}%
    \begin{minipage}{\dimexpr\linewidth-2\fboxrule\relax}
      \setlength{\fboxsep}{6pt}%
      \noindent\colorbox{promptaccent!9}{%
        \parbox{\dimexpr\linewidth-2\fboxsep\relax}{%
          \raggedright\fontsize{9}{10.5}\selectfont\bfseries
          \color{promptaccent!75!black}#2}}\par
      \nointerlineskip
      {\color{promptaccent!35}\hrule height 0.3pt}%
      \vspace{6pt}%
}{%
      \vspace{6pt}%
    \end{minipage}%
  \end{lrbox}%
  \noindent\fcolorbox{promptaccent!55}{promptaccent!2}{\usebox{\joltpromptbox}}%
  \par\endgroup\addvspace{8pt}%
}

\label{app:prompts-lcb}

Each teacher context receives a generation that passed every training-visible test, if one exists, and otherwise the execution feedback for the corresponding attempt. Feedback can include a failing input, expected
and observed output, or execution error. The student receives the following user message with no additional system message.

\begin{joltprompt}[joltstudent]{Student User Message}
\begin{lstlisting}[style=joltprompt]
You are a coding expert. You will be given a coding problem, and you need to write a correct Python program that matches the specification and passes all tests. The time limit is 1 second. You may start by outlining your thought process. In the end, please provide the complete code in a code block enclosed with ``` ```.

{problem}
\end{lstlisting}
\end{joltprompt}
\begin{joltprompt}{Addition When Starter Code Specifies a Function Signature}
\begin{lstlisting}[style=joltprompt]
Your solution should have the following signature: ```python
{signature}
```
\end{lstlisting}
\end{joltprompt}
The teacher receives the same user message, a system message, and one of
the two evidence blocks below. The evidence is appended to the user message
after a blank line.

\begin{joltprompt}[joltteacher]{Teacher System Message}
\begin{lstlisting}[style=joltprompt]
You are a coding expert solving a problem with privileged evidence from sibling attempts. Work through the algorithm and edge cases carefully. Return one complete Python solution under the original response contract.
\end{lstlisting}
\end{joltprompt}
\begin{joltprompt}[joltteacher]{Teacher Evidence: Successful Sibling}
\begin{lstlisting}[style=joltprompt]
The following sibling response passed every public test. Use it as evidence while independently solving the original problem, but do not assume that public-test success proves it will pass the held-out tests. Check the algorithm and edge cases, return one complete Python program under the original response contract, and do not mention the sibling response in your answer.

<successful_sibling_response>
{successful_response}
</successful_sibling_response>
\end{lstlisting}
\end{joltprompt}
\begin{joltprompt}[joltteacher]{Teacher Evidence: Execution Feedback}
\begin{lstlisting}[style=joltprompt]
The following is authoritative feedback from evaluating an earlier attempt on only the public test subset. Use it to correctly solve the original problem, but do not assume that passing these tests proves the solution will pass the held-out tests. Check the algorithm and edge cases, and return one complete Python program under the original response contract. Do not mention the feedback in your answer.

<public_test_feedback>
{public_test_feedback}
</public_test_feedback>
\end{lstlisting}
\end{joltprompt}

\endgroup

\FloatBarrier
\subsection{AppWorld}
\suppressfloats[t]
\label{app:appworld-candidates}

\paragraph{Data and evaluation.}
We adapt the AppWorld setup of \citet{wang2026skillsd},
training on all 90 publicly available training tasks.
All reported AppWorld evaluations use the 57 tasks in the
official development split. Training runs for 300 optimizer
updates, with evaluation every 25 updates. At each evaluated
checkpoint, we generate one unprivileged student trajectory
per task and training seed using greedy decoding
(temperature 0), a total context limit of 65,536 tokens,
and a maximum of 50 interaction turns. Evaluation uses
the shared scaffold without retrieved skills. We report
state-test pass rate, defined as the fraction of state tests
passed for each task, averaged across tasks.

\FloatBarrier
\begingroup
\lstdefinestyle{joltprompt}{
  language={},
  basicstyle=\fontsize{8.5}{10.2}\fontencoding{T1}\fontfamily{lmtt}\selectfont,
  columns=fullflexible,
  keepspaces=true,
  breaklines=true,
  breakatwhitespace=true,
  breakindent=0pt,
  breakautoindent=false,
  showstringspaces=false,
  showspaces=false,
  showtabs=false,
  upquote=true,
  tabsize=2,
  frame=none,
  aboveskip=0pt,
  belowskip=0pt,
  xleftmargin=7pt,
  xrightmargin=7pt,
  resetmargins=true
}
\ifdefined\joltpromptbox\else\newsavebox{\joltpromptbox}\fi
\newenvironment{joltprompt}[2][black]{%
  \par\addvspace{7pt}%
  \begingroup
  \colorlet{promptaccent}{#1}%
  \setlength{\fboxsep}{0pt}%
  \setlength{\fboxrule}{0.45pt}%
  \begin{lrbox}{\joltpromptbox}%
    \begin{minipage}{\dimexpr\linewidth-2\fboxrule\relax}
      \setlength{\fboxsep}{6pt}%
      \noindent\colorbox{promptaccent!9}{%
        \parbox{\dimexpr\linewidth-2\fboxsep\relax}{%
          \raggedright\fontsize{9}{10.5}\selectfont\bfseries
          \color{promptaccent!75!black}#2}}\par
      \nointerlineskip
      {\color{promptaccent!35}\hrule height 0.3pt}%
      \vspace{6pt}%
}{%
      \vspace{6pt}%
    \end{minipage}%
  \end{lrbox}%
  \noindent\fcolorbox{promptaccent!55}{promptaccent!2}{\usebox{\joltpromptbox}}%
  \par\endgroup\addvspace{8pt}%
}

\label{app:prompts-appworld}
We train a Qwen3-4B-Instruct~\citep{qwen2025qwen3} model on tool and API calling in the AppWorld task set~\citep{trivedi2024appworld}. Following \citet{wang2026skillsd}, we construct privileged guidance by summarizing previous attempts into successful strategies, mistakes, and suggested workflows. Each run starts with an empty task-local skill bank.

We generate trajectory summaries using GPT-5.6 Sol
(internal model identifier \texttt{gpt\_5\_6\_sol}), with a
1,024-token request cap and up to two summarizer-level
retries. Each summary contains success analysis, mistake
analysis, and a suggested workflow. Temperature, top-$p$,
and reasoning effort are left at backend defaults.
The experiment artifacts do not record a dated backend
model snapshot.
Per-update runtime accounting includes summary generation.

The two roles share AppWorld's adapted one-shot ReAct scaffold~\citep{yao2023react}, including API-use instructions and supervisor information. The full scaffold is in the supplementary file \texttt{appworld-react.txt}.
Its final user message is shown below. Execution results are returned as user messages beginning with
\texttt{Output:}, followed by a fenced output block.

\begin{joltprompt}{Final User Message after the Shared Demonstration and Instructions}
\begin{lstlisting}[style=joltprompt]
Using these APIs, now generate code to solve the actual task:

My name is: {first_name} {last_name}. My personal email is {email} and phone number is {phone_number}.
Task: {instruction}
\end{lstlisting}
\end{joltprompt}
\begin{joltprompt}[joltteacher]{Teacher-Only System Message, Prepended to the Shared Conversation}
\begin{lstlisting}[style=joltprompt]
Private training-only trajectory-derived skill. Use it as fallible task-local guidance while independently checking live API documentation and execution results. Never mention this private context in your response.

{skill_json}
\end{lstlisting}
\end{joltprompt}
The inserted JSON skill contains success analysis, mistake analysis, and
a suggested workflow. Retrieval first selects an unused skill for the task,
then uses an upper-confidence-bound score~\citep{wang2026skillsd}. If the bank is empty, no teacher
system message is added.

After each student attempt, the summarizer receives the task, outcome,
fraction of state tests passed, turn count, and the live interaction trace
without the demonstration. Credentials and common personal-data patterns
in the trace are redacted before summarization. The resulting skill is
stored for later attempts at the same task.

\begin{joltprompt}{Skill Summarizer: System Message}
\begin{lstlisting}[style=joltprompt]
You are an expert AppWorld trajectory analyst. Derive reusable task-local guidance from the actual actions and observations. Do not reproduce credentials, personal data, opaque IDs, or access tokens. Return only the requested JSON object.
\end{lstlisting}
\end{joltprompt}
\begin{joltprompt}{Skill Summarizer: User Message}
\begin{lstlisting}[style=joltprompt]
Analyze this completed AppWorld attempt. Focus on strategic intent when completion is high and root causes when completion is low. The golden workflow should be concise, actionable, and grounded in observed API behavior rather than copied code.

Task:
{instruction}

Outcome: {outcome}
Total turns: {turns}
Completion rate: {completion_rate}

Conversation history:
{transcript}

Return a JSON object containing exactly three nonempty string fields: success_analysis, mistake_analysis, and golden_workflow.
\end{lstlisting}
\end{joltprompt}
\endgroup

\FloatBarrier
\subsection{Terminal-Bench}
\label{app:terminal-details}
\suppressfloats[t]

The terminal setup uses Qwen3-30B-A3B-Instruct~\citep{qwen2025qwen3} on TMax training tasks~\citep{ivison2026tmax},
with a separate stateful shell sandbox for each role. The teacher alone sees
private verifier source and related files in its system context. These files
are not supplied to the student's context, filesystem, or evaluation.
Reward requires an explicit submission marker followed by hidden-verifier
grading.

\paragraph{Evaluation.}
For each trained method, we evaluate the final checkpoint after 50 training
updates on the complete Terminal-Bench 2.1 task
set~\citep{merrill2026terminalbench}. We use the
mini-SWE-agent-based \texttt{Vanillux2Agent} harness with a persistent shell
released by \citet{ivison2026tmax}.\footnote{Source for evaluation defaults:
\url{https://github.com/hamishivi/tmax}, commit \texttt{6d3d606}.}
Following its evaluation defaults, we use temperature 0.7, top-$p=0.95$,
a limit of 16,384 generated tokens per assistant turn, and at most 64 agent
steps. Bash commands have a 120-second timeout; overall trial timeouts follow
Terminal-Bench's per-task defaults without overrides.
Evaluation uses the unprivileged student and omits the teacher-only
verifier information. For each task, we sample 128 trajectories and estimate
pass@$k$ for $k\in\{1,4,16,32\}$ using the estimator of
\citet{chen2021evaluating}, averaging the resulting estimates across tasks.
Due to resource constraints, Table~\ref{tab:terminal-evaluation} reports
results from one training run per method. The 128 evaluation trajectories
per task estimate each checkpoint's pass@$k$; variability across training
seeds is not assessed.

\begingroup
\lstdefinestyle{joltprompt}{
  language={},
  basicstyle=\fontsize{8.5}{10.2}\fontencoding{T1}\fontfamily{lmtt}\selectfont,
  columns=fullflexible,
  keepspaces=true,
  breaklines=true,
  breakatwhitespace=true,
  breakindent=0pt,
  breakautoindent=false,
  showstringspaces=false,
  showspaces=false,
  showtabs=false,
  upquote=true,
  tabsize=2,
  frame=none,
  aboveskip=0pt,
  belowskip=0pt,
  xleftmargin=7pt,
  xrightmargin=7pt,
  resetmargins=true
}
\ifdefined\joltpromptbox\else\newsavebox{\joltpromptbox}\fi
\newenvironment{joltprompt}[2][black]{%
  \par\addvspace{7pt}%
  \begingroup
  \colorlet{promptaccent}{#1}%
  \setlength{\fboxsep}{0pt}%
  \setlength{\fboxrule}{0.45pt}%
  \begin{lrbox}{\joltpromptbox}%
    \begin{minipage}{\dimexpr\linewidth-2\fboxrule\relax}
      \setlength{\fboxsep}{6pt}%
      \noindent\colorbox{promptaccent!9}{%
        \parbox{\dimexpr\linewidth-2\fboxsep\relax}{%
          \raggedright\fontsize{9}{10.5}\selectfont\bfseries
          \color{promptaccent!75!black}#2}}\par
      \nointerlineskip
      {\color{promptaccent!35}\hrule height 0.3pt}%
      \vspace{6pt}%
}{%
      \vspace{6pt}%
    \end{minipage}%
  \end{lrbox}%
  \noindent\fcolorbox{promptaccent!55}{promptaccent!2}{\usebox{\joltpromptbox}}%
  \par\endgroup\addvspace{8pt}%
}

\label{app:prompts-terminal}
The following templates document the current TMax training setup. Each
role receives a system message, the task and interaction instructions, and
a \texttt{bash} tool accepting one string argument, \texttt{command}.
Braced names denote substituted fields. The complete shared user prompt
and JSON tool schema are supplied in the supplementary materials; the
user-prompt opening and submission rule are reproduced below.

\begin{joltprompt}{System Message (Both Roles)}
\begin{lstlisting}[style=joltprompt]
You are a helpful assistant that can interact with a computer.

Your response must include a THOUGHT section before your action where you explain your reasoning. After the THOUGHT, you must call the `bash` tool with EXACTLY ONE bash command (multiple commands chained with `&&` or `||` count as a single action).

Failure to follow these rules - calling no tool, calling a tool other than `bash`, or omitting the THOUGHT - will cause your response to be rejected.
\end{lstlisting}
\end{joltprompt}
\begin{joltprompt}{Shared User Prompt: Opening}
\begin{lstlisting}[style=joltprompt]
Please solve this task:

{instruction}

You can execute bash commands and edit files (with `sed`, `cat > file << 'EOF'`, etc.) to implement the necessary changes.
\end{lstlisting}
\end{joltprompt}
\begin{joltprompt}{Shared User Prompt: Submission Instruction}
\begin{lstlisting}[style=joltprompt]
6. Submit your changes and finish your work by issuing the following command:
   `echo COMPLETE_TASK_AND_SUBMIT_FINAL_OUTPUT`
   Do not combine it with any other command. After this command, you cannot continue working on the task.
\end{lstlisting}
\end{joltprompt}
The remaining shared instructions describe inspecting the environment,
implementing and testing changes, persistent shell state, noninteractive
commands, and truncated output. The tool definition describes a stateful
working directory and exported environment variables.

\begin{joltprompt}[joltteacher]{Teacher-Only Addition to the System Message}
\begin{lstlisting}[style=joltprompt]
You are the privileged teacher during training. The private verifier files below define the success criterion. Use them to derive a correct solution, but do not attempt to edit `/tests` or rely on the verifier being present in the sandbox: it is uploaded only after your work is complete.

<private_verifier>
{verifier_files}
</private_verifier>
\end{lstlisting}
\end{joltprompt}
The verifier block concatenates the supplied files in order, with each
file rendered as \texttt{--- \{path\} ---} followed by its contents on a
new line and a blank line between files. Terminal-Bench evaluation omits
this teacher-only block.
\endgroup

\FloatBarrier
\subsection{Consistency Across Settings}
\label{app:comparison-uncertainty}

Table~\ref{tab:evaluation-summary} reports mean performance and standard
deviations across training runs. To complement these summaries, we compare
JOLT+ with GRPO using the average probability of improvement proposed by
\citet{agarwal2021statistical}. For each setting, we compare all pairs of
final-checkpoint evaluation runs, recording whether the JOLT+ score exceeds
the GRPO score and assigning half credit to ties. We then average these
probabilities, giving equal weight to GSM8K, MATH, LCBv6, and AppWorld.
This metric summarizes the consistency of the observed ordering across runs.

The analysis uses the individual evaluation scores shown in
Figures~\ref{fig:accuracy-summary-kl-grid-preview}
and~\ref{fig:evaluation-summary-grid}.
We quantify uncertainty in this aggregate probability using a stratified
percentile bootstrap,
independently resampling training runs with replacement within each method
and setting while preserving the run counts and keeping the four settings
fixed. We evaluate the empirical bootstrap distribution exactly. We resample
training runs because the pairwise comparisons share observations.

As shown in Table~\ref{tab:probability-improvement}, JOLT+ outperforms GRPO
in nearly all observed run comparisons. Every observed JOLT+ run exceeds
every observed GRPO run on LCBv6 and AppWorld. The average probability of
improvement is $94.7\%$, with a nominal 95\% percentile bootstrap interval
of $[83.3\%,100.0\%]$. These results provide consistent empirical support
for JOLT+ across the evaluated settings. With only three runs in most
settings, bootstrap intervals can underestimate uncertainty, particularly
when all observed comparisons favor one method~\citep{agarwal2021statistical}.
Terminal-Bench has one training run per method and is reported separately
in Table~\ref{tab:terminal-evaluation}.

\begin{table}[!htbp]
  \centering
  \small
  \caption{Empirical probability that a JOLT+ run outperforms a GRPO run
  at the final checkpoint, using the evaluations in
  Figures~\ref{fig:accuracy-summary-kl-grid-preview}
  and~\ref{fig:evaluation-summary-grid}. The average gives equal weight
  to each setting.}
  \label{tab:probability-improvement}
  \setlength{\tabcolsep}{8pt}
  \renewcommand{\arraystretch}{1.1}
  \begin{tabular}{@{}lccr@{}}
    \toprule
    Setting & JOLT+ Runs & GRPO Runs & Probability of Improvement (\%) \\
    \midrule
    GSM8K & 5 & 4 & 90.0 \\
    MATH & 3 & 3 & 88.9 \\
    LCBv6 & 3 & 3 & 100.0 \\
    AppWorld & 3 & 3 & 100.0 \\
    \midrule
    \textbf{Equal-weight average} & --- & --- & \textbf{94.7} \\
    \bottomrule
  \end{tabular}
\end{table}

\FloatBarrier

\section{Token Efficiency and Scaling}
\suppressfloats[t]
\label{app:token-details}

We study how additional training budget translates into student accuracy on
MATH. Our experiments combine token-budget sweeps with checkpoints from
extended training runs, covering both early learning and performance at
larger budgets. Figure~\ref{fig:math-token-efficiency} summarizes this
relationship using completion tokens and processed training tokens as
measures of training cost.

\paragraph{Experimental setup.}
We vary learning rates and training budgets for each method and measure
performance at checkpoints throughout training. The budget sweeps
characterize the accuracy attainable at different costs, while the extended
runs examine whether additional training yields further improvement.
Figure~\ref{fig:math-token-efficiency} shows each method's Pareto frontier
based on training-token expenditure and mean evaluation accuracy. Token counts reflect actual
expenditure, including any overshoot of the target budget.

\paragraph{Token accounting.}
Completion tokens count student completions for GRPO and OPSD, and both
teacher and student completions for JOLT. Processed training tokens count
unpadded prompt and completion positions processed during optimization and
cross-context scoring, including teacher-only context tokens, scoring passes
without gradient computation, and repeated optimization passes. The MATH
experiments use worked solutions supplied by the dataset, with no additional
hint generation, retry rollouts, or discarded trajectories. These metrics
measure token expenditure and do not directly quantify wall-clock savings.

\paragraph{Curve fitting.}
Following \citet{khatri2025art}, we fit a sigmoidal relationship between
accuracy and log training-token budget:
\begin{equation}
  a(B)=a_0+\frac{a_\infty-a_0}{1+(B_{50}/B)^k},
  \label{eq:token-scaling-fit}
\end{equation}
where $B$ is the budget in millions of tokens and $a_0$ is the measured
base-model accuracy. The parameter $a_\infty$ describes the fitted plateau,
$B_{50}$ is the budget required to reach half the fitted accuracy gain, and
$k$ controls the slope with respect to log budget.

We estimate the three free parameters separately for each method and token
counter using unweighted bounded nonlinear least squares. We use multiple
initializations and retain the solution with the smallest residual sum of
squares. Table~\ref{tab:token-scaling-coefficients} reports the fitted
coefficients and root-mean-squared error (RMSE).

\begin{table}[!htbp]
\centering
\small
\caption{Fitted accuracy--token scaling parameters for
Figure~\ref{fig:math-token-efficiency}. Budgets are in millions of tokens;
RMSE is in percentage points.}
\label{tab:token-scaling-coefficients}
\begin{tabular}{llrrrr}
\toprule
Token counter & Method & $a_\infty$ (\%) & $B_{50}$ & $k$ & RMSE \\
\midrule
Completion & GRPO & 70.044 & 4.173 & 1.655 & 0.994 \\
           & OPSD & 61.706 & 0.512 & 2.076 & 0.938 \\
           & JOLT & 72.123 & 0.440 & 2.413 & 1.126 \\
\midrule
Processed  & GRPO & 70.441 & 4.343 & 1.501 & 0.959 \\
           & OPSD & 61.691 & 1.314 & 2.270 & 0.937 \\
           & JOLT & 72.125 & 0.955 & 2.634 & 1.152 \\
\bottomrule
\end{tabular}
\end{table}

\paragraph{Efficiency comparison.}
We compare methods at the GRPO checkpoint near the knee of its learning
curve, where mean evaluation accuracy is $63.44\%$. For each token counter,
we read GRPO's budget from this checkpoint in
Figure~\ref{fig:math-token-efficiency} and obtain JOLT's budget at the same
accuracy by inverting its fitted curve, using the measured base accuracy
$a_0\simeq43.85\%$. Table~\ref{tab:token-efficiency-comparison} reports the
resulting budgets and their ratios. These are approximate point estimates
from the mean-accuracy comparison; confidence intervals for the budget
ratios are not estimated.

\begin{table}[!htbp]
\centering
\small
\caption{Token budgets at $63.44\%$ MATH accuracy. GRPO uses the observed
checkpoint near its learning-curve knee; JOLT uses its fitted curve.
Budgets are in millions of tokens, and values are approximate.}
\label{tab:token-efficiency-comparison}
\begin{tabular}{lrrr}
\toprule
Token counter & GRPO & JOLT & GRPO/JOLT \\
\midrule
Completion & 8.04 & 0.617 & $13.0\times$ \\
Processed & 8.40 & 1.30 & $6.5\times$ \\
\bottomrule
\end{tabular}
\end{table}

\paragraph{Sensitivity to target accuracy.}
To examine how the estimated token savings depend on the comparison target,
we evaluate both methods' fitted curves at $55\%$, $60\%$, and $65\%$ MATH
accuracy. These targets lie within the observed accuracy ranges of both
methods. We invert each fitted curve to estimate the required token budget,
\begin{equation}
  B(a)=B_{50}\left(\frac{a-a_0}{a_\infty-a}\right)^{1/k},
\end{equation}
and report the ratio of GRPO's budget to JOLT's in
Table~\ref{tab:token-target-sensitivity}. Across these targets, JOLT requires
approximately $9.5$--$14.4\times$ fewer completion tokens and
$4.3$--$7.4\times$ fewer processed training tokens. The estimated advantage
therefore persists across this accuracy range, although its magnitude depends
on the target. These comparisons use fitted budgets for both methods and are
point estimates without confidence intervals.

\begin{table}[!htbp]
\centering
\small
\caption{Estimated token budgets and GRPO/JOLT budget ratios at multiple
MATH accuracy targets. Both methods' budgets are obtained from their fitted
curves. Budgets are in millions of tokens; ratios are calculated before
rounding.}
\label{tab:token-target-sensitivity}
\begin{tabular}{crrrrrr}
\toprule
Target & \multicolumn{3}{c}{Completion tokens}
       & \multicolumn{3}{c}{Processed training tokens} \\
\cmidrule(lr){2-4}\cmidrule(lr){5-7}
accuracy & GRPO & JOLT & Ratio & GRPO & JOLT & Ratio \\
\midrule
$55\%$ & 3.48 & 0.368 & $9.5\times$  & 3.50  & 0.811 & $4.3\times$ \\
$60\%$ & 5.56 & 0.496 & $11.2\times$ & 5.81  & 1.065 & $5.5\times$ \\
$65\%$ & 9.92 & 0.691 & $14.4\times$ & 10.73 & 1.443 & $7.4\times$ \\
\bottomrule
\end{tabular}
\end{table}

\FloatBarrier
\section{Teacher-Objective Ablations}
\suppressfloats[t]
\label{app:teacher-ablation-details}
\label{sec:math_teacher_ablations}

\begin{figure}[!tb]
\centering
\includegraphics[width=\linewidth]{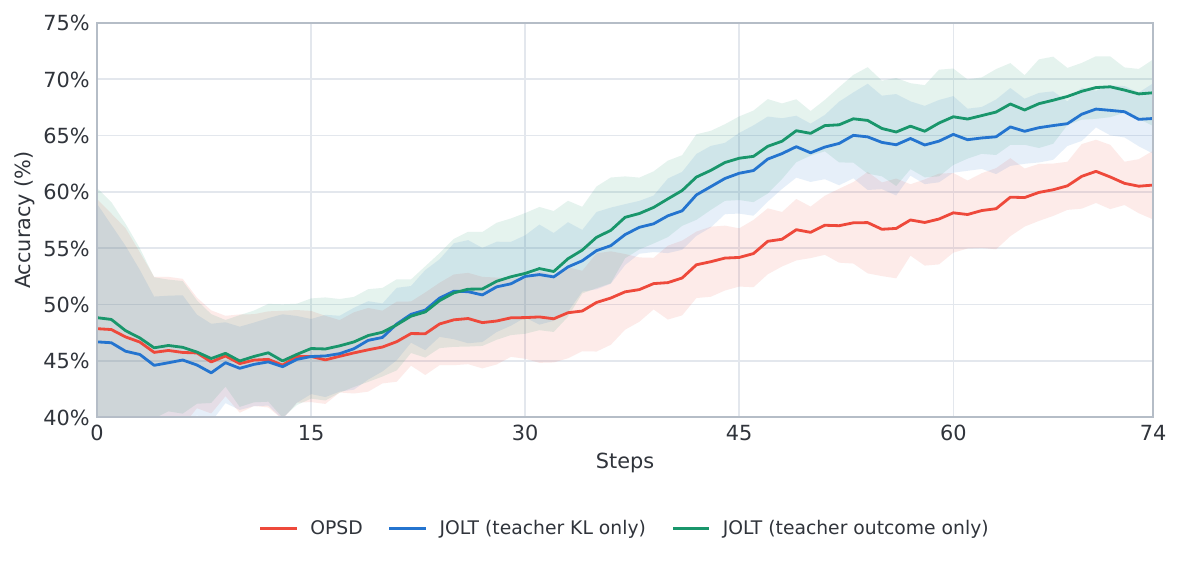}
\par\vspace{-4pt}
\caption{Learning curves on MATH for OPSD and JOLT variants with either
teacher KL regularization or teacher outcome rewards. Both teacher-training
variants achieve higher accuracy than OPSD later in training. Lines and
shaded regions show the mean and 95\% confidence intervals across five
training seeds.}
\label{fig:teacher-objective-ablation}
\end{figure}

We investigate how explicit teacher optimization affects student learning,
with student distillation retained in every configuration.
We compare OPSD with two JOLT variants that retain only the teacher KL term
or only the teacher outcome-reward term. OPSD updates the shared model through
student distillation, while the JOLT variants additionally optimize the model
under the privileged teacher context. These comparisons examine whether
either teacher-objective component can improve student learning.
We additionally evaluate OPSD with student outcome rewards, implemented by
removing both teacher-objective terms from JOLT+ while retaining its student
learning configuration.
Figure~\ref{fig:teacher-objective-ablation} shows the teacher-objective
ablation curves, and
Table~\ref{tab:math_opsd_teacher_objectives} reports the selected
configurations and held-out accuracy, including JOLT and JOLT+ at step 75.

\paragraph{Experimental setup and tuning.}
We use the same model, task setup, and evaluation protocol as our MATH
experiment. For OPSD and the two teacher-objective variants, we tune each
configuration independently, searching learning
rates of $\{10^{-6},2\times10^{-6},4\times10^{-6}\}$, with additional rates
of $\{2.5,5,7.5\}\times10^{-7}$ for the teacher KL-only variant.
Table~\ref{tab:math_opsd_teacher_objectives} reports the selected learning
rates and teacher-objective coefficients.
The OPSD + Student Reward control uses the same student loss weights,
normalization, rollout count, and training schedule as JOLT+, with no
explicit teacher optimization. Its privileged teacher still shares the
student's parameters and therefore changes through student updates.

\begin{table}[!htbp]
\centering
\small
\setlength{\tabcolsep}{3pt}
\begin{tabularx}{\linewidth}{@{}>{\raggedright\arraybackslash}X c c c c c@{}}
\toprule
Configuration
& \shortstack[c]{Student outcome\\reward}
& \shortstack[c]{Teacher outcome\\coefficient $\alpha$}
& \shortstack[c]{Teacher KL\\coefficient $\beta$}
& Tuned LR
& \shortstack[c]{Held-out\\accuracy (\%)} \\
\midrule
OPSD & -- & -- & -- & $2\times10^{-6}$ & $62.95\pm0.78$ \\
OPSD + Student Reward
& $\checkmark$ & -- & -- & $2\times10^{-6}$ & $66.52\pm0.73$ \\
\midrule
JOLT (teacher KL only)
& -- & -- & $10^{-6}$ & $10^{-6}$ & $65.39\pm0.97$ \\
JOLT (teacher outcome only)
& -- & $1$ & -- & $2\times10^{-6}$ & $67.87\pm0.86$ \\
\midrule
JOLT (full objective)
& -- & $1$ & $10^{-3}$ & $2\times10^{-6}$ & $72.10\pm0.82$ \\
JOLT+ (full objective)
& $\checkmark$ & $1$ & $10^{-3}$ & $2\times10^{-6}$ & $73.80\pm0.74$ \\
\bottomrule
\end{tabularx}
\caption{Held-out MATH accuracy at step 75. OPSD, OPSD + Student Reward,
and the two teacher-objective variants report means across five shared
training seeds, with 95\% confidence-interval half-widths.
OPSD + Student Reward disables both teacher-objective terms in JOLT+.
JOLT and JOLT+ use the step-75 evaluations in
Figure~\ref{fig:evaluation-summary-grid}, each with three runs and a 95\%
confidence interval; their configurations are given in
Table~\ref{tab:hyperparameter-summary}. JOLT+'s interval is a Student-$t$
interval computed from the plotted standard deviation.
Dashes indicate absent objective terms.}
\label{tab:math_opsd_teacher_objectives}
\end{table}

\paragraph{Student Rewards without Teacher Optimization.}
This control combines sparse student rewards with dense privileged
self-distillation, as also studied by \citet{penaloza2026privileged},
without applying either teacher-objective term. At step 75, it attains
$66.52\pm0.73\%$ held-out accuracy, compared with $73.80\pm0.74\%$ for
JOLT+. With the student learning configuration held fixed, this comparison
supports a benefit from explicit teacher optimization beyond combining
student rewards with distillation on MATH. Full JOLT also achieves higher
accuracy than this control without using student outcome rewards.

\paragraph{Effect of Teacher Optimization.}
As shown in Figure~\ref{fig:teacher-objective-ablation}, both teacher-training
variants achieve higher accuracy than OPSD as training progresses. These
gains also appear in the held-out evaluations in
Table~\ref{tab:math_opsd_teacher_objectives}, where the teacher outcome-only
variant achieves higher accuracy than the KL-only variant. The improvement
from the KL-only variant suggests that
encouraging agreement with the student can provide a useful teacher-training
signal even without teacher outcome rewards. Together, these results
support explicitly optimizing the teacher as a way to improve student
learning. Because both roles share parameters, these comparisons do not
separate improvements in the distillation signal from direct transfer
through teacher updates.

\FloatBarrier
\section{Token-Level Alignment}
\label{app:token-alignment}

For each student-generated token, we compare teacher and student log
probabilities conditioned on the same response prefix. Red indicates a
higher log probability under the teacher, while blue indicates a lower
log probability under the teacher. Color intensity reflects the magnitude
of this difference, using symmetric limits around zero shared across all
methods within each figure. The visualizations use each method's final
checkpoint, with the displayed problem and responses selected at random.

\begin{figure}[!t]
  \centering
  \setlength{\abovecaptionskip}{4pt}
  \includegraphics[width=\linewidth]{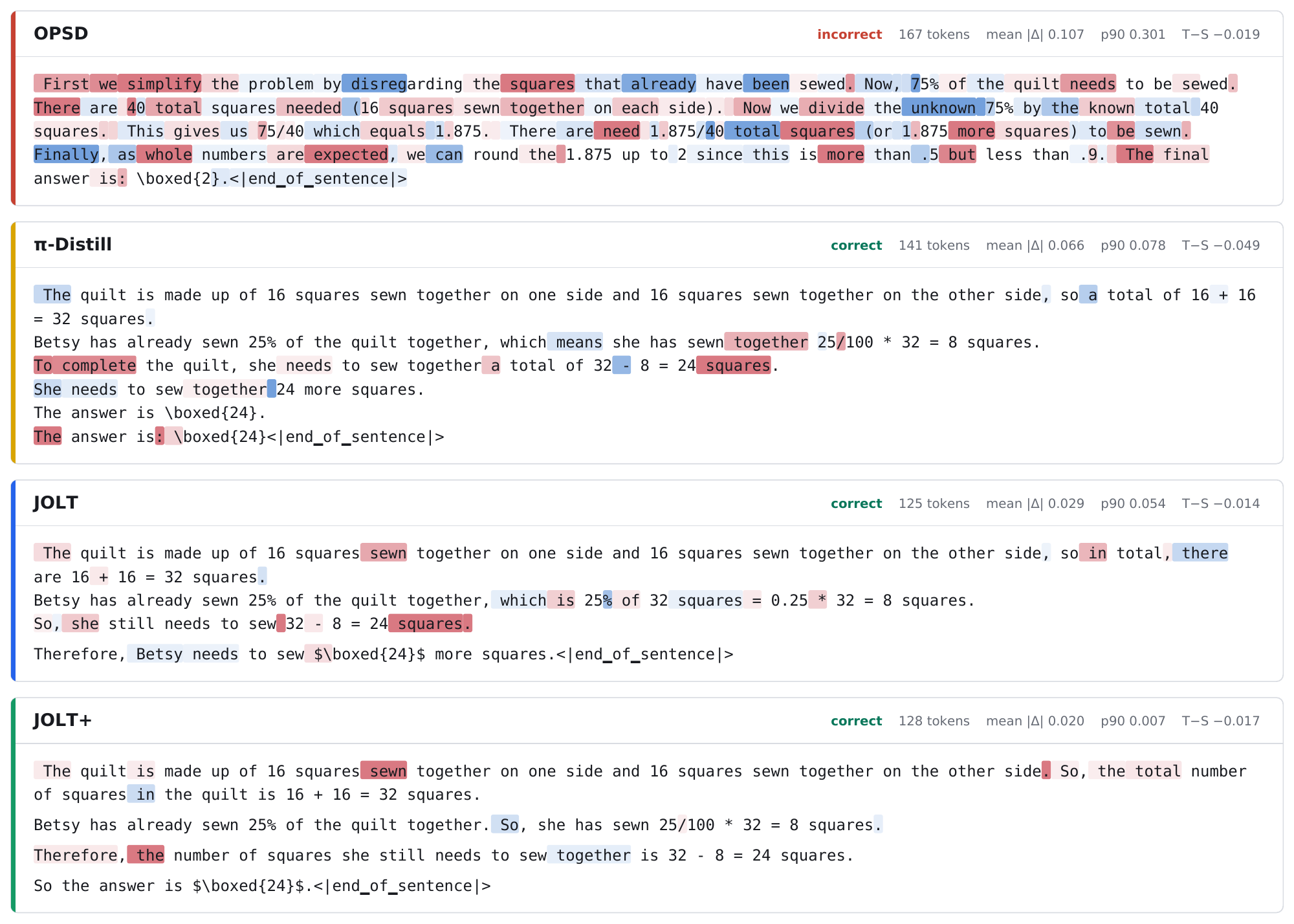}
  \par
  \caption{Token-level teacher--student alignment on the same problem.
  Red indicates higher teacher log probability, while blue indicates lower
  teacher log probability.
  JOLT and JOLT+ show smaller cross-context log-probability differences than
  OPSD and $\pi$-Distill.}
  \label{fig:token-logprob-all-methods}
\end{figure}

Figure~\ref{fig:token-logprob-all-methods} extends the token-level comparison in
Figure~\ref{fig:token-logprob-example} to include $\pi$-Distill and JOLT+.
JOLT and JOLT+ exhibit smaller teacher--student log-probability differences
throughout their generations, consistent with the closer contextual agreement
observed during training.

\FloatBarrier
\begingroup
\setlength{\parskip}{4pt}
\setlength{\textfloatsep}{12pt plus 2pt minus 2pt}
\section{Gradient Bias--Variance Analysis}
\suppressfloats[t]
\label{app:gradient-details}

We characterize the student learning signals encountered during training
through empirical bias and variance measurements.
Figure~\ref{fig:gradient-student-overlay} reports these statistics at
checkpoints along each method's own training trajectory. We complement
these statistics with comparisons of normalized gradient directions at
fixed checkpoints.

\paragraph{Reference and measurements.}
The expected reward gradient is not directly available for these models. We
construct a Monte Carlo reference by averaging eight independent REINFORCE
gradient estimates~\citep{williams1992simple}, each using 512 completions per
problem. The reference uses group-centered returns and
token normalization and serves as an empirical comparison target rather than
an exact expected-reward gradient. We evaluate eight batches of eight MATH
problems at initialization and training steps 1, 2, 5, 10, 20, 35, 50, and 75.
For each batch, we hold the model weights and problems fixed and compute
$K=32$ independent gradient estimates, each using eight student completions
per problem. We measure their bias and variance relative to the reference.

Gradients are computed with respect to all model parameters, with
privileged-teacher probabilities treated as fixed targets. The distillation
gradients use the student's reverse-KL objective at temperature $1.1$.
Each loss uses a global token average: token-level contributions are summed
across all sampled completions in the problem batch and divided by the total
number of valid completion tokens. Gradients are measured before
gradient-norm clipping.
The measured student gradients are $0.1g_D$ for JOLT and
$0.1g_D+g_R$ for JOLT+, where $g_D$ and $g_R$ denote the distillation
and REINFORCE gradients. The two components use the same sampled student
completions, preserving their covariance in the combined estimate.

These coefficients affect the scale of the reported statistics. Multiplying
a gradient estimator by $\alpha>0$ scales its variance by $\alpha^2$ without
changing its direction, while its Euclidean discrepancy from a fixed
reference also depends on $\alpha$. Consequently, lower measured variance or
discrepancy alone does not establish stronger directional alignment.

Let $g_1,\ldots,g_K$ denote the sampled gradient estimates,
$g_{\mathrm{ref}}$ the reference gradient, and
$\bar g=K^{-1}\sum_{k=1}^{K}g_k$ the mean sampled gradient. We measure empirical
squared bias and variance as
\begin{equation}
 \widehat B^2=\|\bar g-g_{\mathrm{ref}}\|^2,
 \qquad
 \widehat V=\frac{1}{K-1}\sum_{k=1}^{K}\|g_k-\bar g\|^2.
\end{equation}
Squared bias measures disagreement between the average estimate and the
reference, while variance measures fluctuations around that average. Their
contributions to mean squared error satisfy
\begin{equation}
 \widehat{\mathrm{MSE}}
 =\frac{1}{K}\sum_{k=1}^{K}\|g_k-g_{\mathrm{ref}}\|^2
 =\widehat B^2+\frac{K-1}{K}\widehat V.
 \label{eq:empirical-mse-decomposition}
\end{equation}
We compute each statistic separately within each problem batch and average
the resulting statistics equally across batches.

Gradient variance is measured across repeated trajectory samples with the
checkpoint and problem batch held fixed. For each method, we analyze the run
with the best historical late-training performance. Confidence bands are
pointwise 95\% intervals from 100,000 paired bootstrap resamples of the eight
problem batches, keeping each batch together across checkpoints and
estimators. The reference is held fixed during resampling, so these intervals
exclude uncertainty across training seeds and reference samples.

These measurements concern student gradients at the policies reached by each
method. Differences therefore reflect both the learned policies and their
update estimators. The Euclidean statistics depend on gradient magnitude and
do not directly test directional reward consistency, which is defined up to
positive scaling. The empirical squared-bias statistic includes sampling
uncertainty in both the estimated gradient mean and the reference. Across the
evaluated batches, checkpoints, and trajectories, the estimated Monte Carlo
variance of the reference mean divided by its squared norm ranges from
$0.074$ to $0.661$. Reference uncertainty therefore contributes to the measured
discrepancy, particularly when the reference gradient is small. Teacher
learning influences the checkpoints being evaluated, but its gradient is
excluded from the measurements. The analysis therefore does not establish
alignment of the combined shared-parameter update.

\paragraph{Comparisons during training.}
Figure~\ref{fig:gradient-student-overlay} evaluates each method's student
gradient along its own training run, constructing a reference at each
checkpoint. GRPO's error is predominantly variance, whereas OPSD exhibits low
variance but increasing disagreement with the reference. JOLT and JOLT+
maintain substantially smaller disagreement than OPSD while retaining low
variance. These comparisons describe the sampled student learning signals
along the evaluated training trajectories.

\begin{figure}[!t]
\centering
\includegraphics[width=\linewidth]{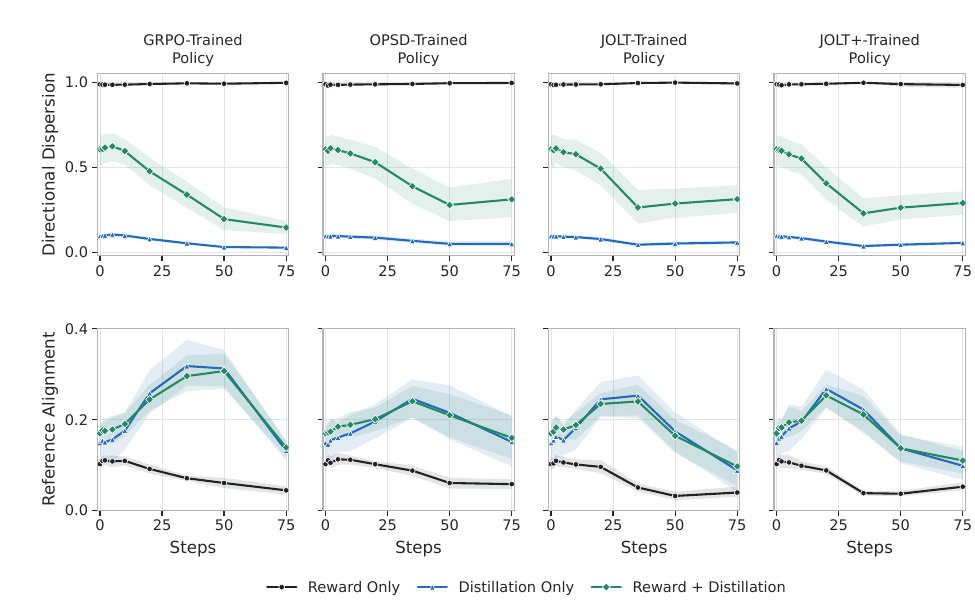}
\caption{Student-gradient directions at fixed checkpoints. Distillation-only
and combined estimates show lower directional dispersion (top) and higher
mean cosine similarity to the Monte Carlo reward-gradient reference (bottom)
than reward-only estimates. Columns follow different training methods;
within each checkpoint, estimators share model weights, problems, sampled
student trajectories, and the reference. Bands show pointwise 95\%
whole-batch bootstrap intervals.}
\label{fig:matched-gradient-updates}
\end{figure}

\paragraph{Directional comparisons at fixed checkpoints.}
Figure~\ref{fig:matched-gradient-updates} compares reward-only,
distillation-only, and combined student estimators. Each column follows
checkpoints from one training method. Within each checkpoint, all three
estimators use the same model weights, problem batches, sampled student
trajectories, and reference. The combined estimator is $0.1g_D+g_R$,
using the reverse-KL distillation gradient defined above.

We normalize each sampled gradient before aggregation:
\begin{equation}
 u_k=\frac{g_k}{\|g_k\|},\qquad
 v=\frac{g_{\mathrm{ref}}}{\|g_{\mathrm{ref}}\|},\qquad
 \bar u=\frac{1}{K}\sum_{k=1}^{K}u_k.
\end{equation}
We retain zero gradients as $u_k=0$. Directional dispersion and mean
reference alignment are
\begin{equation}
 D=\frac{1}{K-1}\sum_{k=1}^{K}\|u_k-\bar u\|^2,\qquad
 P=\frac{1}{K}\sum_{k=1}^{K}\langle u_k,v\rangle.
 \label{eq:direction-diagnostic}
\end{equation}
Lower $D$ indicates more consistent directions across sampled trajectories;
$P$ is the mean per-estimate cosine similarity to the reference, with zero
gradients contributing zero. We compute these statistics within each problem
batch and average equally across batches, using the same bootstrap procedure
described above.

At the evaluated checkpoints, distillation-only and combined estimates have
lower directional dispersion and higher mean reference alignment than
reward-only estimates. These differences persist after normalizing gradient
magnitude and therefore cannot be explained solely by an overall rescaling
of the gradients. The pattern appears across all four trained policies.
These comparisons characterize student estimators at fixed checkpoints;
they do not isolate the effect of teacher training or measure reward
improvement from the complete joint update. Uncertainty in the Monte Carlo
reference also affects the directional measurements.
\par
\endgroup

\FloatBarrier
\Needspace{9\baselineskip}
\section{\texorpdfstring{$\pi$}{pi}-Distill: Hyperparameter Sensitivity}
\label{app:pi-distill-sensitivity}

To better understand $\pi$-Distill's uneven performance across tasks, we
investigate its sensitivity to the learning rate, teacher--student mixing
weight, and teacher KL coefficient. We evaluate selected configurations
across multiple random seeds and, on MATH, additionally examine
learning-rate schedules, mixing warmup, and separate normalization of the
teacher and student losses. Table~\ref{tab:pi-distill-search-coverage}
summarizes the search coverage for each setting.

\begin{table}[H]
\centering
\small
\caption{Search coverage for $\pi$-Distill hyperparameter tuning.}
\label{tab:pi-distill-search-coverage}
\renewcommand{\arraystretch}{1.15}
\begin{tabularx}{\linewidth}{@{}l>{\raggedright\arraybackslash}X@{}}
\toprule
Setting & Search Coverage \\
\midrule
GSM8K & Five learning rates evaluated over the full training horizon, followed
by repeated-seed comparisons. \\
MATH & We examine 42 configurations spanning learning rates, teacher–student mixing weights, teacher KL coefficients, and scheduling choices, including teacher-only controls. Selected configurations are evaluated across additional random seeds. \\
LCBv6 & Eighteen distinct configurations spanning eight learning rates, five
teacher KL coefficients, and three teacher mixing weights, followed by
fresh-seed confirmation. \\
AppWorld & Five full-horizon learning-rate trials and additional replications
of the selected configuration. \\
\bottomrule
\end{tabularx}
\end{table}

The weaker MATH performance persists across these choices. Larger learning
rates and stronger student replay often aggravate degradation, while
schedule and normalization changes do not consistently restore sustained
improvement. Some configurations that appear promising during initial
screening also fail to reproduce their gains across additional seeds. These
observations suggest that the difficulty extends beyond selecting an
appropriate update scale or balancing the two roles. The behavior nevertheless varies across settings: on AppWorld,
$\pi$-Distill achieves a development-set state-test pass rate close to GRPO
and substantially above OPSD.

\FloatBarrier

\end{document}